\documentclass[onecolumn,11pt]{article}
\usepackage{pgf}
\usepackage{amsmath}
\usepackage{amssymb}
\usepackage{subfigure}
\usepackage{multirow}
\usepackage[latin9]{inputenc}
\usepackage{breakurl}
\usepackage{pgf}
\usepackage{subfigure}
\usepackage{multirow}
\usepackage{bbm}
\usepackage{cite}
\usepackage{float}
\usepackage{bbold}
\usepackage{algorithm}
\usepackage[noend]{algpseudocode}
\usepackage{graphicx}

\iffalse
\newcommand{\todo}[1]{{\textcolor{red}{[[TODO: {#1}]]}}}

\newcommand{\highlight}[1]{{\textcolor{blue}{{#1}}}}
\newcommand{\commentfoot}[1]{\footnote{\textcolor{red}{\emph{Comment: #1}}}}
\newcommand{\topic}[1]{}
\else
\newcommand{\todo}[1]{}
\newcommand{\commenttext}[1]{}
\newcommand{\commentfoot}[1]{}
\newcommand{\highlight}[1]{{{#1}}}
\newcommand{\topic}[1]{}
\fi

\iffalse
        \newcommand{\cutsectionup}{\vspace*{-0.1in}}

        \newcommand{\cutsubsectionup}{\vspace*{-0.09in}}

        \newcommand{\cutequationup}{\vspace*{-0.12in}}
        \newcommand{\cutequationdown}{\vspace*{-0.12in}}

\else 
        \newcommand{\cutsectionup}{}

        \newcommand{\cutsubsectionup}{}

        \newcommand{\cutequationup}{}
        \newcommand{\cutequationdown}{}

\fi
\makeatother

\title{Adaptive Questionnaires for Direct Identification of Optimal Product Design}
\usepackage{authblk}
\author[1]{Max Yi Ren\thanks{yiren@asu.edu}}
\author[2]{Clayton Scott\thanks{clayscot@umich.edu}}
\affil[1]{Department of Mechanical Engineering, Arizona State University}
\affil[2]{Division of Electrical and Computer Engineering, University of Michigan, Ann Arbor}

\graphicspath{ {figures/} }

\begin{document}
\maketitle
\begin{abstract}
We consider the problem of identifying the most profitable product design from a finite set of candidates under unknown consumer preference. A standard approach to this problem follows a two-step strategy: First, estimate the preference of the consumer population, represented as a point in part-worth space, using an adaptive discrete-choice questionnaire. Second, integrate the estimated part-worth vector with engineering feasibility and cost models to determine the optimal design. In this work, we (1) demonstrate that accurate preference estimation is neither necessary nor sufficient for identifying the optimal design, (2) introduce a novel adaptive questionnaire that leverages knowledge about engineering feasibility and manufacturing costs to directly determine the optimal design, and (3) interpret product design in terms of a nonlinear segmentation of part-worth space, and use this interpretation to illuminate the intrinsic difficulty of optimal design in the presence of noisy questionnaire responses. We establish the superiority of the proposed approach using a well-documented optimal product design task. This study demonstrates how the identification of optimal product design can be accelerated by integrating marketing and manufacturing knowledge into the adaptive questionnaire.
\end{abstract}

\section{Introduction}
\label{sec:intro}
Understanding consumer preferences is usually necessary during profit-driven product design. A common tool for this cause is Discrete Choice Analysis (DCA). In DCA, the designer collects consumer responses on product alternatives through a questionnaire, and uses the data to create a preference model, which is  {\it subsequently} used for the identification of the optimal (i.e., most profitable) design. While numerous research has followed this two-step procedure (see, e.g., \cite{wassenaar2003approach, michalek2005linking, wassenaar2005enhancing,lewis2006decision, michalek2011enhancing,namwoo2014charging}), \highlight{there has been limited theoretical discussion of the relationship between the accuracy of a preference model and the market performance of the resultant product (Some empirical studies can be found in \cite{long2015should,shinexploring})}. \highlight{Although} having a precise understanding of consumer preference does not hurt, it may not be necessary for product design: For example, when all but one candidate designs have higher cost than their potential profit, that single design is optimal regardless of our understanding of consumer preference. While this is an extreme case, it suggests that we could potentially accomplish the task of identifying the optimal design with fewer resources spent on the questionnaire if it directly identifies the optimal design without focusing on preference learning as an intermediate step.

This paper aims to turn this intuition into a rigorous questionnaire mechanism. 
To accomplish this goal, we argue that finding the optimal product is a {\it group identification} problem (defined below), where the {\it group label} (the ID of the optimal design) of an {\it object} (the consumer preference model) is to be identified through binary {\it queries} (the questionnaire). We further show that an existing algorithm for group identification can be extended to optimal product design. 

By drawing this connection, we propose a questionnaire mechanism that adaptively chooses queries to greedily minimize the expected number of queries. We compare this approach with a standard adaptive questionnaire by Abernethy et al.~\cite{abernethy2008eliciting} (called Abernethy's alg. in the sequel) on a well-documented product design problem. The performance of the two methods reflects their different objectives: While Abernethy's alg. yields better preference estimates, our method has consistently better performance in identifying the optimal product, leading to a significant difference in the expected profit of the resultant product.    

The key contributions of this paper are as follows. (1) We clearly explain the mathematical difference between preference modeling and optimal product identification, and thus show that accurate preference estimation is neither necessary nor sufficient for identifying the optimal product design. To summarize, consider the true preference as a point in a ``preference space'': The former is equivalent to estimating the location of that point. In the latter, the entire space is segmented with each segment representing preferences that share the same optimal design. The objective of the latter is thus to identify the true segment, i.e., the segment where the true preference lies. (2) We develop a real-time adaptive questionnaire mechanism that identifies optimal designs more effectively than a standard method in noise-free scenarios, by leveraging engineering feasibility and cost models. (3) We point out a critical issue {\it by construction} in the use of preference models in product design: Even when the estimation of preference is close to the truth, the model may still lead to an inferior design decision. This happens when the true segment is ``thin'' or close to the origin, with the latter case representing low sensitivity of consumer preferences to design changes or equivalently, high noise in responses to choice questions. The results presented in this paper call for a more systematic product design methodology when its goal is to serve product designers to identify profitable designs more cost effectively rather than to help marketing researchers and psychologists to precisely understand human preference.

The rest of the paper is structured as follows. Sec.~\ref{sec:prelim} provides preliminary knowledge necessary for in-depth discussion. Sec.~\ref{sec:act} introduces the questionnaire mechanism and Sec.~\ref{sec:implementation} its implementation. Sec.~\ref{sec:case} presents results from a case study. Lastly, discussions on the limitations of the proposed method and relaxations of its assumptions are presented in Sec.~\ref{sec:disc}.

\section{Preliminaries}
\label{sec:prelim}
We introduce \highlight{background material} necessary for an elaboration on this problem statement.
\subsection{Profit-driven product design}
\label{sec:productdesign}
A profit-driven product design problem can be formulated as follows, with ${\bf z}$ being the product design and ${\bf w}$ the parameters of a preference model: 
\begin{equation}
\max_{{\bf z}\in\mathcal{Z}} ~\text{profit}({\bf z};{\bf w}):= \text{market share}({\bf z};{\bf w})\left(\text{price}({\bf z}) - \text{cost}({\bf z})\right).
\label{eq:marketobj}
\end{equation}
More specifically, ${\bf z}:=(z_{(1)},\cdots,z_{(D)})\in \mathcal{Z}$ is an binary attribute vector where each element represents the existence of a certain attribute level in the product\footnote{\highlight{For example, the attribute ``MPG'' of a car has levels ``20'', ``30'', and ``40''. The binary encoding of attributes we use in this paper, e.g., $z_{(1)=1}\Leftrightarrow \text{MPG}=20$, $z_{(2)=1}\Leftrightarrow \text{MPG}=30$, etc., is a common way to treat nonlinearity in preference with respect to attribute levels.}}. $\mathcal{Z}$ represents the set of all feasible combinations of attribute levels. Unit price is indicated by one of the attributes while unit cost is a function of ${\bf z}$ often built upon engineering models of the problem. Parameters ${\bf w} \in \mathbb{R}^D$ are called {\it part-worths} of a consumer {\it utility} model $u({\bf z};{\bf w})$. The preference influences the profit through the market share. Specifically, we use the logit model~\cite{mcfadden1973conditional}
\begin{equation}
\text{market share}({\bf z};{\bf w}) \propto \text{sigmoid}(u({\bf z};{\bf w})-u({\bf z}_0;{\bf w})),
\label{eq:choice}
\end{equation} 
where ${\bf z}_0$ is some fixed competing product to ${\bf z}$. Extension to multiple competing products is trivial. The RHS of the equation represents the probability of choosing design ${\bf z}$ by a consumer, and is proportional to the market share under homogeneous consumer preference, i.e., everyone shares the same deterministic ${\bf w}$. 
As discussed in Sec.~\ref{sec:disc}, other choice models, e.g., probit and non-compensatory ones~\cite{hauser2010disjunctions,morrow2014market,shin2015modeling}, can replace Eq.~\eqref{eq:choice} without significantly affecting our discussion.

\cutsubsectionup
\subsection{Preference learning}
\label{sec:preferencelearning}
Clearly, ${\bf w}$ plays a critical role in formulating and solving Eq.~\eqref{eq:marketobj}. In reality, however, the true part-worths of a target consumer group, denoted as ${\bf w}^*$, are often unknown. DCA derives a part-worth estimate, denoted as $\hat{\bf w}$, using a series of queries. Each query consists of a set of design alternatives, from which the participant is asked to pick the relatively more preferred design. The accumulated responses form a data set $\mathcal{S}^{(Q)}:=\{({\bf z}_{q_1},{\bf z}_{q_2},\cdots)\}_{q=1}^{Q}$ where ${\bf z}_{q_1}$ is the preferred design in the $q$th query, and $\{{\bf z}_{q_2},\cdots\}$ are the unchosen alternatives. In the sequel, we will assume pairwise choices, and discuss the extension to multiple choices in Sec.~\ref{sec:disc}. When necessary, superscripts will be omitted to avoid clutter. Given this data set and a prior belief of consumer preference $p({\bf w})$, the posterior density function $p({\bf w};\mathcal{S}^{(Q)})$ can be derived following a regularized logistic or hierarchical Bayes~\cite{lenk1996hierarchical,rossi2003bayesian} model. Under either model, the maximum a posteriori (MAP) estimator can be found as $\hat{\bf w}:=\text{argmax}_{{\bf w}} p({\bf w};\mathcal{S})$, and the uncertainty of this point estimate, i.e., its variance-covariance matrix, can be calculated by the inverse of the Hessian ${\bf H}(\hat{\bf w},\mathcal{S})$ of the negative log-likelihood $-\log\left(p({\bf w};\mathcal{S})\right)$. In this paper, we use a regularized logistic model due to its convexity and real-time computation of MAP estimates. A comparison between these two approaches is discussed in \cite{toubia200712}. \highlight{Assuming a regularized logistic model, we have:
\begin{equation}
-\log\left(p({\bf w};\mathcal{S})\right) = \sum_{q'=1}^{q}\log (1+{\bf w}^T \Delta {\bf z}^{(q')}) + \frac{C}{2}{\bf w}^T{\bf w},
\label{eq:nll}
\end{equation}
where $\Delta {\bf z}^{(q')}:= {\bf z}_{q'_2} - {\bf z}_{q'_1}$, and the parameter $C$ will be selected by cross-validation (see Sec.~\ref{sec:implementation}).} The goal of preference learning is to find a part-worth estimate close to the ground truth and with low uncertainty (e.g., determinant of ${\bf H}(\hat{\bf w},\mathcal{S})$), while limiting $Q$, the number of queries posed.

\cutsubsectionup
\subsection{Optimal product identification}
\label{sec:optid}
Since the uncertainty in part-worth estimation always exists under a finite number of queries, one needs to modify Eq.~\eqref{eq:marketobj} to incorporate $p({\bf w};\mathcal{S})$. Two modified problem definitions for optimal product identification exist: One is to optimize the expected profit:
\begin{equation}
    \max_{{\bf z}\in\mathcal{Z}} ~\mathbb{E}_{{\bf w}}\left[\text{profit}({\bf z};{\bf w})\right] = \int_{\mathbb{R}^D} \text{profit}({\bf z};{\bf w})p({\bf w};\mathcal{S})
    d{\bf w},
\label{eq:marketobj1}
\end{equation}
and the second is to optimize the probability of being the most profitable (denoted as $\pi({\bf z})$):
\begin{equation}
\begin{aligned}
\max_{{\bf z}\in\mathcal{Z}} ~\pi({\bf z}) := \mathbb{E}_{{\bf w}}\left[\mathbbm{1}({\bf z};{\bf w})\right]
= \int_{\mathbb{R}^D} \mathbbm{1}({\bf z};{\bf w})p({\bf w};\mathcal{S})
    d{\bf w},
\end{aligned}
\label{eq:marketobj2}
\end{equation}
where $\mathbbm{1}({\bf z};{\bf w}):=\mathbbm{1}\left(\text{profit}({\bf z};{\bf w})>\text{profit}({\bf z}';{\bf w}), \forall {\bf z}'\neq {\bf z} \right)$ and $\mathbbm{1}(\text{condition})=1$ when the condition is true and $0$ otherwise. The design decision generated from these two objectives may not be consistent for an arbitrary $\mathcal{S}$. The goal of optimal product identification is to acquire such queries and responses so that the design decision matches the true solution to Eq.~\eqref{eq:marketobj}.

\cutsubsectionup
\subsection{Assumptions}
\label{sec:assumption}
The majority of the discussion will be based on the following conditions and assumptions: (1) We consider a consumer group who share similar preferences; (2) queries are binary, i.e., each consists of two design alternatives and one must be chosen; (3) $\mathcal{Z}$ is a finite set; (4) the preference model is linear, i.e., $u = {\bf w}^T{\bf z}$; (5) $||{\bf w}^*||$ is a large number, reflecting low-noise responses and substantial information for measurement; (6) engineering models, i.e., product costs and feasibility, are deterministic; and (7) the predicted optimal design is derived from Eq.~\eqref{eq:marketobj2}. We will discuss in Subsec.~\ref{subsec:noise} the issue with response noise; and in Sec.~\ref{sec:disc} the relaxation of assumptions (1) to (4), as well as the connection between Eqs.~\eqref{eq:marketobj1} and \eqref{eq:marketobj2}.

\subsection{Query strategies for preference learning and optimal product identification}
\label{sec:motivation}
A query strategy defines how the $q$th query shall be created based on the current knowledge $p({\bf w};\mathcal{S}^{(q-1)})$. For preference learning purposes, existing strategies~\cite{toubia2004polyhedral,toubia2003fast, toubia200712,abernethy2008eliciting} have been developed to greedily minimize the estimation uncertainty of $\hat{\bf w}$, which is equivalent to maximizing the determinant of ${\bf H}(\hat{\bf w},\mathcal{S})$ (see Subsec.~\ref{subsec:benchmark} for a discussion of \cite{abernethy2008eliciting}). Below we show that the query strategies for optimal product identification should be different than for preference learning. Without loss of generality, we will use a simple 2D case for visualization. 

\begin{figure}
\centering
\includegraphics[width=\linewidth]{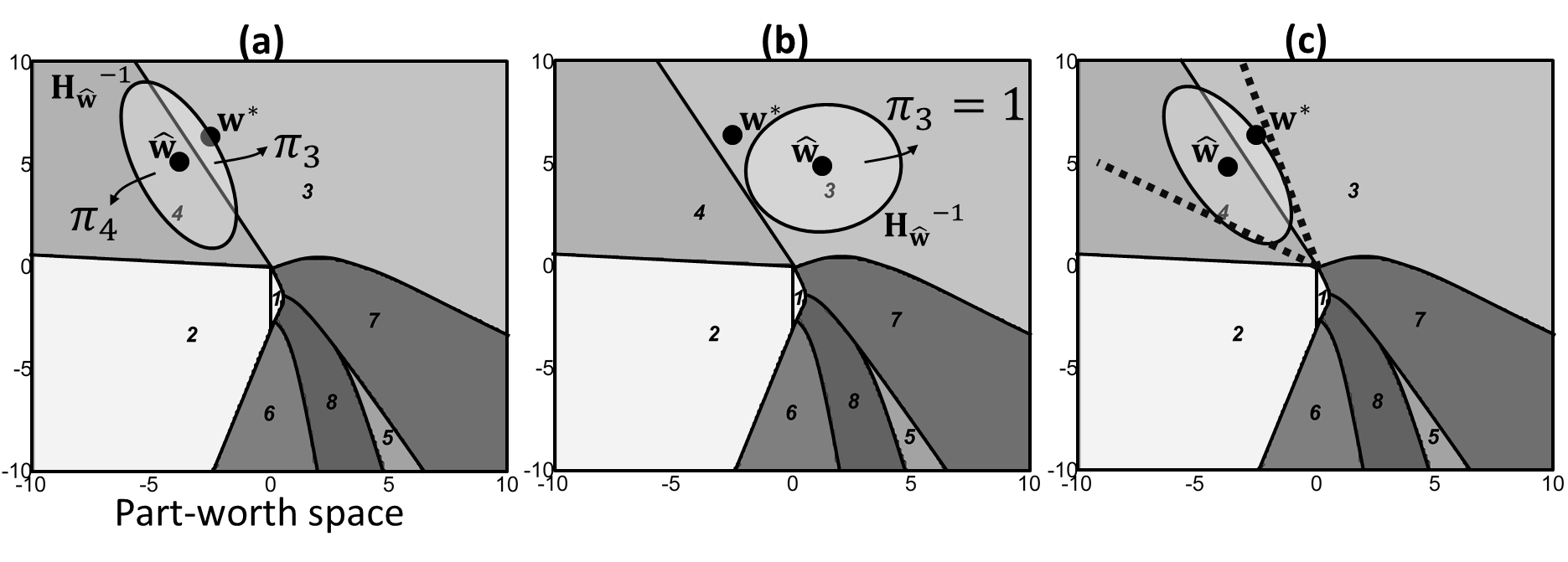}
\caption{(a) A good estimation $\hat{\bf w} \approx {\bf w}^*$ does not guarantee $\hat{\bf z} = {\bf z}_{k^*}$. \highlight{$\pi_3$ and $\pi_4$ are the probabilities for product 3 and 4 to be the optimal, respectively.} (b) Achieving $\hat{\bf z} = {\bf z}_{k^*}$ does not necessarily require $\hat{\bf w} \approx {\bf w}^*$. (c) Each query can be considered as a cut in the part-worth space. The posterior $p({\bf w};\mathcal{S})$ is shifted to the convex cone defined by these cuts. The uncertainty in $\hat{\bf w}$ is illustrated by the covariance ellipse ${\bf H}_{\hat{\bf w}}^{-1}$. Generation of the figure: In the space of $[-10,10]^2$, we sample eight points uniformly as candidate products, i.e., each one is represented by two continuous product attributes, one of which is the price. For each part-worth on a dense grid of the space, we calculate ${\bf z}_{k^*}$ by Eq.~\eqref{eq:marketobj2}. Grid points with the same ${\bf z}_{k^*}$ share the same color.} 
\label{fig:2dsegment}
\end{figure}

Consider $K$ candidate products with fixed costs. By setting the true part-worth to some arbitrary ${\bf w}$, we can solve Eq.~\eqref{eq:marketobj} to derive the optimal product for that particular ${\bf w}$. Doing so for all ${\bf w} \in \mathbb{R}^2$ leads to a segmentation of the part-worth space, as illustrated in Fig.~\ref{fig:2dsegment}. Each segment, denoted as $\mathcal{W}_k$, corresponds to a set of ${\bf w}$ that share the same most profitable design ${\bf z}_k$. Due to the nonlinearity in the mapping from preference to profit (see Eqs.~\eqref{eq:marketobj} and \eqref{eq:choice}), the segmentation is also nonlinear. To elaborate, any two candidate products ${\bf z}_1$ and ${\bf z}_2$ have the same profit when ${\bf w}$ satisfies 
\begin{equation}
\frac{\exp({\bf w}^T{\bf z}_1)(\text{price}({\bf z}_1)-\text{cost}({\bf z}_1))}{\exp({\bf w}^T{\bf z}_1)+\exp({\bf w}^T{\bf z}_0)} = \frac{\exp({\bf w}^T{\bf z}_2)(\text{price}({\bf z}_2)-\text{cost}({\bf z}_2))}{\exp({\bf w}^T{\bf z}_2)+\exp({\bf w}^T{\bf z}_0)}.
\end{equation}

Let the true optimal design, induced by ${\bf w}^*$, be ${\bf z}_{k^*}$, and the solution to Eq.~\eqref{eq:marketobj2} be $\hat{\bf z}$. Ideally, one would like to obtain a perfect preference model ($\hat{\bf w} = {\bf w}^*$) and thus the correct optimal product ($\hat{\bf z} = {\bf z}_{k^*}$). However, we show in Figs.~\ref{fig:2dsegment}a-b that a good estimation $\hat{\bf w} \approx {\bf w}^*$ does not guarantee $\hat{\bf z} = {\bf z}_{k^*}$, neither does $\hat{\bf z} = {\bf z}_{k^*}$ require $\hat{\bf w} \approx {\bf w}^*$. In addition, readers may notice that the segments cluster around the origin, in which region consumers are less sensitive to design changes\footnote{Low part-worth values can also be interpreted as high response noise. Salisbury and Feinberg~\cite{salisbury2010alleviating} and Louviere et al.~\cite{louviere2002dissecting} pointed out the downsides of mixing these two interpretations. But we do not discuss this issue here.}. As we will discuss in Subsec.~\ref{sec:results}, this clustering causes difficulty in correctly identifying the optimal design under indifferent preference or noisy responses. It is worth noting, however, that due to the indifference or high noise in consumer preference, all segments near the origin lead to similar expected profit as market shares are more evenly distributed among competing products.

Knowing this, we can further discuss the difference between query strategies for preference learning and for optimal product identification. We start from a prior probability density $p({\bf w})$ (e.g., a multivariate normal distribution or a distribution that leads to maximum entropy, see the entropy definition in Subsec.~\ref{subsec:group}). When a user response $({\bf z}_{q_1}, {\bf z}_{q_2})$ is collected from iteration $q$, we update $p({\bf w};\mathcal{S}^{(q-1)})$ so that the probability mass is shifted towards the half space $\{{\bf w}| {\bf w}^T\left({\bf z}_{q_1}-{\bf
z}_{q_2}\right)>0\}$. After some $Q$ responses are collected, $p({\bf
w}|\mathcal{S}^{(Q)})$ will concentrate in a cone that contains ${\bf w}^*$ (Fig.~\ref{fig:2dsegment}c). A good query strategy for learning ${\bf w}^*$ is thus to
quickly ``narrow'' this cone. 
The algorithms proposed by Toubia et al.~\cite{toubia2004polyhedral,toubia2003fast}, Abernethy et al.~\cite{abernethy2008eliciting}, Tong and Koller~\cite{tong2002support}, and Jamieson and Nowak~\cite{jamieson2011active} all follow the idea of narrowing the feasible part-worth space by halving it. 

\begin{figure}
\centering
\includegraphics[width=\linewidth]{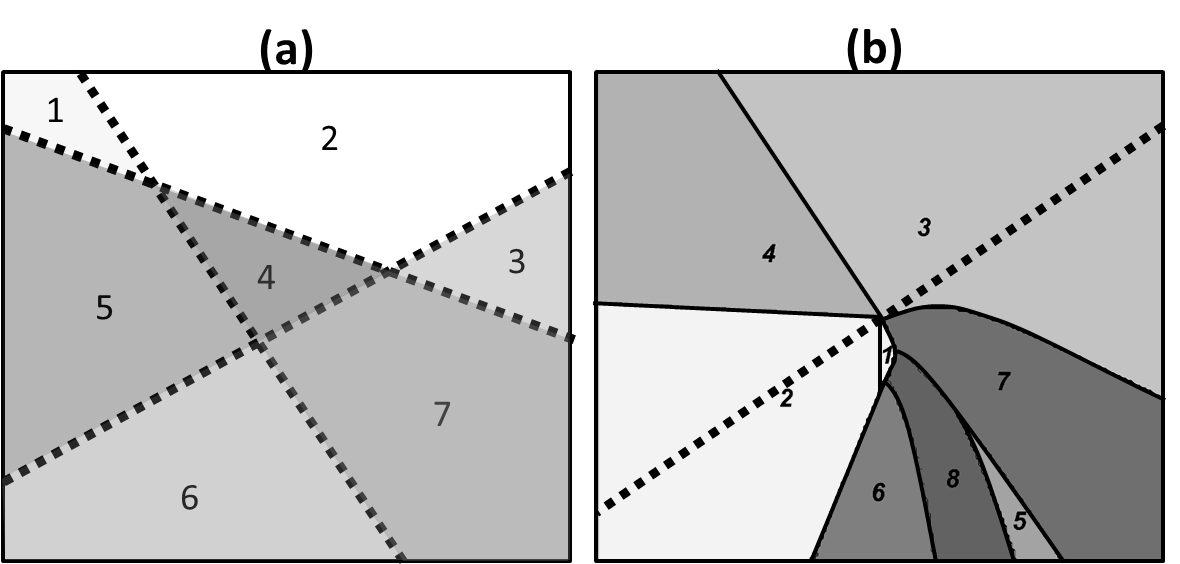}
\caption{Comparison on problem settings between (a) \cite{jamieson2011active}  and (b) this paper: (a) Each segment (shaded polygons) represents a preference ranking and each hyperplane a pair-wise comparison query (dotted lines) (b) The nonlinear segments (shaded) are induced by the candidate products and their costs, while queries (dotted line) are pair-wise comparisons} 
\label{fig:compare}
\end{figure}

For optimal product identification, on the other hand, ${\bf z}_{k^*}$ is correctly identified when $p({\bf w};\mathcal{S}^{(Q)})$ concentrates in the segment $\mathcal{W}_{k^*}$, i.e., when $p({\bf w};\mathcal{S})>0$ only in $\mathcal{W}_k$, $\hat{\bf z} = {\bf z}_{k^*}$ according to both Eq.~\eqref{eq:marketobj1} and Eq.~\eqref{eq:marketobj2}. Therefore, a good query strategy in this case should be to quickly identify the correct ``group'' $\mathcal{W}_k$, by shifting the mass of $p({\bf w};\mathcal{S}^{(Q)})$ towards it, rather than towards the point ${\bf w}^*$. Jamieson and Nowak~\cite{jamieson2011active} shares a similar problem to ours in that a particular segment rather than a point is to be identified from a segmentation of a space. However, in \cite{jamieson2011active}, the segments are created by hyperplanes directly defined by pair-wise comparisons, while in this paper, the nonlinear segments are induced from the candidate products, independent from the hyperplanes defined by the collection of queries. See Fig.~\ref{fig:compare} for a comparison.

\cutsectionup
\section{Query Strategy}
\label{sec:act}
This section addresses the central question of how queries should be adaptively
made to directly minimize the expected number of queries needed to identify the optimal design. To do so, we show the connection between this problem and the problem of ``group identification", and introduce an extension to the Group Identification Splitting Algorithm~\cite{bellala2012group} that solves our problem. We also review an existing strategy for preference learning from Abernethy et al.~\cite{abernethy2008eliciting} (called Abernethy's alg. in the sequel) that will be used as a benchmark in the case study. Implementation details of GISA and Abernethy's alg. will be discussed in Sec.~\ref{sec:implementation}.

\cutsubsectionup
\subsection{Group identification and optimal product identification}
\label{subsec:group}
{\it Group identification} is an extension of {\it object identification}, also known
as the twenty-questions game. Consider a game between players A and B. Both
players are familiar with a list of objects and the group labels they belong to, see Table
\ref{tb:table2}. Player A first picks an object from the list randomly. In object identification, Player B needs to identify the object selected by Player A, whereas in group identification, Player B only needs to determine the group to which the object belongs. To know which group Player A picked from, Player B can pick queries in the columns to ask. Based on Player A's responses, which are binary, Player B either picks new queries or makes a guess on the group label. Player B's query strategy affects how quickly he can identify the correct group. In this example, an intuitively good query to start with is ``Q1'', since its answer directly determines which group the object is from, i.e., it allows one of the group labels to have a probability of one conditioned on the response. It is worth noting that we cannot consider groups as ``meta-objects'', since objects in the same group can lead to different query responses. 
\begin{table}
\centering
\caption{An example of the group identification problem} 
{\footnotesize
\begin{tabular}{l|cccc}
\hline
 Group & \multicolumn{2}{c}{Heroes} & \multicolumn{2}{c}{Villains} \\
 Object & Superman & Batman & Catwoman & Joker\\
 Prior Prob. & 0.25 & 0.25 & 0.25 & 0.25 \\
 \hline
 Q1:Mask? & Yes & Yes & No & No \\
 Q2:Can fly? & Yes & No & No & No \\
 Q3:Female? & No & No & Yes & No \\
\hline
\end{tabular}
}
\label{tb:table2}
\end{table}

We now show that optimal product identification in our context shares the same elements as in
Table~\ref{tb:table2}. Consider the set of objects to be all part-worth vectors in
$\mathbb{R}^D$. Each induces a list of binary responses to pairwise queries, which are defined by $\mathcal{Z}$. Specifically, queries are determined by $\text{sign}\{{\bf w}^T({\bf z}_i-{\bf z}_j)\}$ for all pairs ${\bf z}_i$, ${\bf z}_j$. A group is
a set $\mathcal{W}_k$ of part-worth vectors that share the same most profitable
design. Similar to Player B who has a (non-informative) prior knowledge of which object Player A picks, here objects are associated with a prior probability density $p({\bf w})$, and each group has a prior probability mass $\pi_k^{(0)} = \int_{\mathcal{W}_k} p({\bf w})d{\bf w}$. Upon acquiring the $q$th user response, we can update $\mathcal{S}^{(q)}$, the posterior $p({\bf w};\mathcal{S}^{(q)})$ and  
\begin{equation}
\pi_k^{(q)} = \int_{\mathcal{W}_k} p({\bf w};\mathcal{S}^{(q)})d{\bf w}.
\label{eq:pik}
\end{equation}
Since $p({\bf w};\mathcal{S}^{(q)})$ is a probability density function, we have $\sum_{k=1}^K\pi_k^{(q)}=1$. One can see that both group identification and optimal product identification share the same process of converging $\pi^{(Q)}_{k^*}$ to $1$ by collecting responses to binary queries. The differences from the latter are that (1) we have an infinite number of objects (candidate part-worth vectors), and (2) the conditional probabilities are not proportional to the prior. For example, in Table \ref{tb:table2}, the conditional probabilities for ``superman'' and ``batman'' are still equal after receiving ``yes'' from Q1; while in the case of optimal product identification, each response will nonlinearly affect $p({\bf w};\mathcal{S}^{(q)})$, \highlight{which in turn determines $\Pi^{(q)} := (\pi_1^{(q)},...,\pi_K^{(q)})$}.


\cutsubsectionup
\subsection{The Group Identification Splitting Algorithm (GISA)}
\label{subsec:alg}
\begin{figure}
\centering
\includegraphics[width=0.8\linewidth]{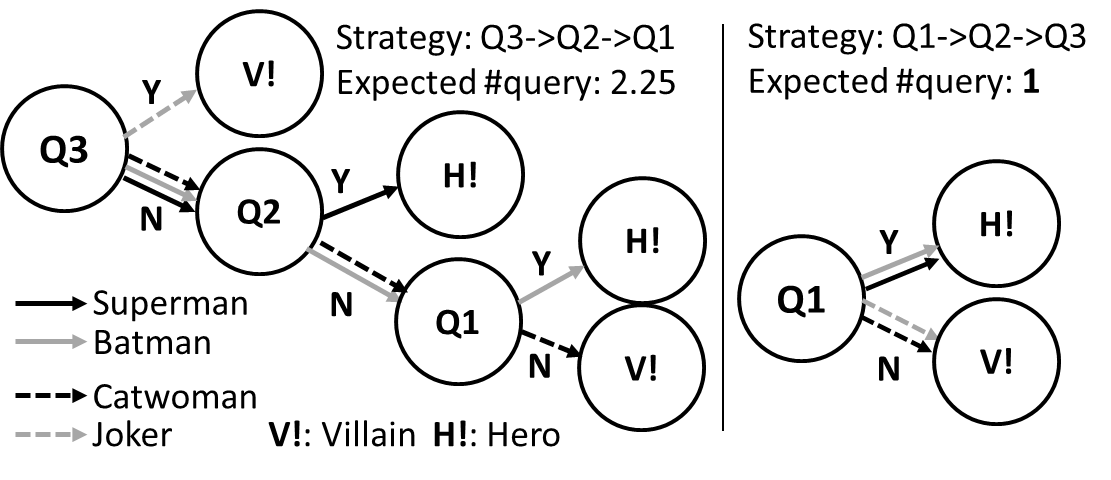}\label{fig:tree}
\caption{Binary decision trees under two query strategies.}
\end{figure}

GISA greedily minimizes the expected number of queries needed to reach $H(\Pi^{(Q)})=0$, which occurs iff all probability mass is on one design. Here we explain the algorithm in the context of optimal product identification. Readers are referred to \cite{bellala2012group} for the derivation of the algorithm for group identification. Let a ``path'' be a sequence of queries and responses. Upon receiving the response for each query, we move from one node on the path to the next, and update $\Pi^{(q)}$. With enough queries, we will reach $\pi^{(Q)}_{k^*} = 1$. Given a query strategy, one can generate multiple paths for objects (${\bf w}$s) drawn from the prior ($p({\bf w})$). These paths form a binary decision tree (Fig.~\ref{fig:tree}). Note that multiple leaf nodes of the tree can share the same group label. The expected number of queries needed to reach a leaf, denoted as $L$, can be
calculated as a function of $p({\bf w})$ and the query strategy.
While optimizing $L$ with respect to all possible query strategies is shown to
be NP complete~\cite{hyafil1976constructing}, Bellala et at. showed that a greedy approach to solve the problem is available as $L$ can be decomposed into a set of additive terms $\tilde{L}^{(a)}$ with respect
to each internal node $a$ of the binary tree~\cite{bellala2012group}.
Therefore, one greedy strategy is to minimize $\tilde{L}^{(q)}$ from node $1$ to $Q$ along the actual path taken during a questionnaire. The local objective $\tilde{L}^{(q)}$ takes the following form:
\cutequationup
\begin{equation}
\tilde{L}^{(q)}=1-H(\rho^{(q)})+\sum_{k=1}^{K}\pi_{k}^{(q)} H(\rho_{k}^{(q)}),
\cutequationdown
\label{eq:obj}
\end{equation}
where $H(\pi):=-\pi \log_2 \pi - (1-\pi)\log_2 (1-\pi)$ is the binary entropy of scalar $\pi$, and $0\log_2 0 :=0$. The ``reduction factor'' $\rho^{(q)}$ and the ``group reduction factor'' $\rho_{k}^{(q)}$ depend on the candidate query, and are defined as follows. For a binary query based on ${\bf z}_{i}$ and ${\bf z}_{j}$, the two child nodes, denoted as $l$ (for left) and $r$ (for right), are induced by responses $({\bf z}_{i}, {\bf z}_{j})$ and $({\bf z}_{j}, {\bf
z}_{i})$, respectively. We define
\begin{equation}
\pi_{l}^{(q)} = \int \mathbbm{1}\left({\bf w}^T({\bf z}_{i}-{\bf
z}_{j})>0\right) p({\bf w};\mathcal{S}^{(q-1)})d{\bf w}
\end{equation}
and $\pi_{r}^{(q)} = 1-\pi_{l}^{(q)}$. These are the conditional probabilities of branching left and right, respectively. $\rho^{(q)}$ is defined as
\cutequationup
\begin{equation}
\rho^{(q)}=\max\{\pi_{l}^{(q)},\pi_{r}^{(q)}\}.
\label{eq_rho}
\cutequationdown
\end{equation}
Similarly, for each design $k$ we define 
\begin{equation}
\pi_{l,k}^{(q)} = \int_{\mathcal{W}_k} \mathbbm{1}\left({\bf w}^T({\bf z}_{i}-{\bf
z}_{j})>0\right) p({\bf w};\mathcal{S}^{(q-1)})d{\bf
w},
\end{equation}
and $\pi_{r,k}^{(q)} = \pi_{k}^{(q)}-\pi_{l,k}^{(q)}$. These are the probabilities of branching left and right, respectively, for part-worths that make ${\bf z}_k$ the optimal. $\rho_{k}^{(q)}$ is defined as
\cutequationup
\begin{equation}
\rho_{k}^{(q)}=\max\{\pi_{l,k}^{(q)},\pi_{r,k}^{(q)}\}/\pi_{k}^{(q)}.
\label{eq_rhok}
\cutequationdown
\end{equation}

GISA picks from the candidate query set $\mathcal{Q}$ a new query that minimizes $\tilde{L}^{(q)}$. An ideal query would achieve $\rho^{(q)}=0.5$ and $\rho_{k}^{(q)}= 0$ or $1$, i.e., the query ``halves'' the part-worth space (by $\rho^{(q)}=0.5$) but preserves entire groups (by $\rho_{k}^{(q)}= 0$ or $1$). In practice, however, the two parts of the objective often trade off. In contrast, algorithms for preference estimation are only concerned with the former criterion. We also terminate the questionnaire after a pre-determined number $Q$ of queries rather than when $H(\Pi^{(q)})=0$. It should be noted that in the original presentation of GISA~\cite{bellala2012group}, the space of objects was discrete and thus calculation of $\pi_k^{(q)}$, $\pi_l^{(q)}$, and $\pi_{l,k}^{(q)}$ was straightforward. Our extension to optimal product identification is thus nontrivial as we must develop computationally efficient methods for calculating these probabilities. 
Alg.~\ref{alg:gisa} provides a sketch of the algorithm. Implementation details will be discussed in Subsec.~\ref{sec:implementationgisa}. 

\begin{algorithm}
\caption{Group Identification Splitting Algorithm}\label{alg:gisa}
\begin{algorithmic}[1]
\State {\bf Input}: $\mathcal{Z}$, $p({\bf w})$, $q=0$, $Q$, unit cost, $u_0$
\State {\bf Output}: $\Pi^{(Q)}$
\While {$q\leq Q$}
\State calculate $\Pi^{(q)}$ (Eq.~\eqref{eq:pik})
\State calculate $\rho^{(q)}$ and $\rho_{k}^{(q)}$ (Eqs.~\eqref{eq_rho}, \eqref{eq_rhok})
\State calculate $\tilde{L}_q$ (Eq.~\eqref{eq:obj}) for each query in $\mathcal{Q}$
\State choose the next query with the minimal $\tilde{L}_q$
\State collect query response, set $q=q+1$, update $\mathcal{S}^{(q)}$ and $p({\bf w};\mathcal{S}^{(q)})$
\EndWhile
\end{algorithmic}
\end{algorithm}

To demonstrate the algorithm, let us revisit the group identification
problem from Table \ref{tb:table2}. For Player B to pick a query, he will first specify prior probability masses for each object to be the one Player A chooses.
Without any knowledge, he assumes that the four objects have equal chances, leading to $\Pi=(\pi_{\text{Heroes}},\pi_{\text{Villains}})=(0.5,0.5)$. The calculation of $\tilde{L}$ follows Eqs.~\eqref{eq:obj}, \eqref{eq_rho} and \eqref{eq_rhok} and is summarized in Table \ref{tb:La}, where $l$ and $r$ represent child nodes for responses ``yes'' and ``no'', respectively. From the table, ``Q1'' has the minimum $\tilde{L}$ and thus will be chosen. It should be noted that the choice of query depends on the prior knowledge. For example, if it is known that Player A will never choose ``Joker'', then Q1 and Q2 are equally plausible.
\begin{table*}[ht!]
\centering
\caption{Query selection for the case in Table \ref{tb:table2}} {\small
\begin{tabular*}{0.9\linewidth}{cccc}
\hline
Query & Q1 & Q2 & Q3 \\
\hline
$\pi_{l}$ $\left(\pi_{r}\right)$ & 0.5 (0.5) & \highlight{0.25 (0.75)} & 0.25 (0.75) \\
$\rho$ & 0.5 & \highlight{0.75} & 0.75 \\
$\pi_{l,\text{(Heroes)}}$ $\left(\pi_{r,\text{(Heroes)}}\right)$ & 1 (0) & 0.5 (0.5) & 0 (1) \\
$\pi_{l,\text{(Villains)}}$ $\left(\pi_{r,\text{(Villains)}}\right)$ & 0 (1) & \highlight{0 (1)} & 0.5 (0.5) \\
$\rho_{\text{(Heroes)}}$ & 1 & 0.5 & 1 \\
$\rho_{\text{(Villains)}}$ & 1 & \highlight{1} & 0.5 \\
$\tilde{L}$ & {\bf 0.5} & \highlight{1.16} & 1.16 \\
\hline
\end{tabular*}
}
\label{tb:La}
\end{table*}

\cutsubsectionup
\subsection{Abernethy's algorithm}
\label{subsec:benchmark}
Abernethy's alg. chooses a new query based on two criteria: (1) ``utility balance" requires the query to have the smallest utility gap $|\left({\bf z}_{i}-{\bf z}_{j}\right)^T\hat{\bf w}|$ where $\hat{\bf w}$ is the current part-worth estimator; and (2) ``minimax variance'' requires the query to minimize the largest estimation variance. For the latter criterion, we calculate the Hessian of the negative log likelihood ${\bf H}(\hat{\bf w},\mathcal{S}^{(q)}):=\frac{\partial^2 \left(-\log p({\bf w};\mathcal{S}^{(q)})\right)}{\partial {\bf w}^2}|_{{\bf w}=\hat{\bf w}}$:
\begin{equation}
{\bf H}(\hat{\bf w},\mathcal{S}^{(q)}) = \frac{1}{C}{\bf I} + \sum_{q'=1}^{q} \frac{\exp(-\hat{\bf w}^T\Delta {\bf z}^{(q')})}{\left(1+\exp(-\hat{\bf w}^T\Delta {\bf z}^{(q')})\right)^2}\Delta {\bf z}^{(q')}\Delta {\bf z}^{(q')T},
\label{eq:hessian}
\end{equation}
and its projection to the subspace perpendicular to $\hat{\bf w}$:
\begin{equation}
{\bf H}_{\hat{\bf w}} = \left({\bf I} - \frac{\hat{\bf w}\hat{\bf
w}^T}{||\hat{\bf w}||^2}\right){\bf H}(\hat{\bf w},\mathcal{S}^{(q)}).
\label{eq:hessianproj}
\end{equation}

The largest estimation variance of $\hat{\bf w}$ (in the subspace perpendicular to $\hat{\bf w}$) corresponds to the eigenvector ${\bf v}$ of the smallest eigenvalue of ${\bf H}_{\hat{\bf w}}$. The projection along $\hat{\bf w}$ is used to ensure that ${\bf v}^T\hat{\bf w}=0$, which satisfies ``utility balance''. Ideally, ${\bf v}$ satisfies both criteria, if there exist ${\bf z}_{i}$ and ${\bf z}_{j}$ such that ${\bf v}={\bf z}_{i}-{\bf z}_{j}$. In practice, we choose a query so that ${\bf z}_{i}-{\bf z}_{j}$ is close to ${\bf v}$ and perpendicular to $\hat{\bf w}$. Alg.~\ref{alg:abernethy} provides a sketch of this algorithm, and implementation details are deferred to Subsec.~\ref{sec:implementationaabernethy}.

\begin{algorithm}
\caption{Abernethy's alg.}\label{alg:abernethy}
\begin{algorithmic}[1]
\State {\bf Input}: set $\mathcal{Z}$, $p({\bf w})$, $q=0$, $Q$
\State {\bf Output}: $\Pi^{(q)}$, $\hat{\bf w}$
\While {$q \leq Q$}
\State calculate ${\bf H}_{\hat{\bf w}}$ (Eq.~\eqref{eq:hessian})
\State find the eigenvector ${\bf v}$ of the smallest eigenvalue of ${\bf H}_{\hat{\bf w}}$
\State find ${\bf z}_{i,j} = \text{argmin}_{\mathcal{Q}}|{\bf v}^T({\bf z}_{i}-{\bf z}_{j})|$
\State collect query response, set $q=q+1$, update $\mathcal{S}^{(q)}$ and $p({\bf w};\mathcal{S}^{(q)})$
\EndWhile
\end{algorithmic}
\end{algorithm}

\cutsectionup
\section{Algorithm Implementation}
\label{sec:implementation}
Some implementation details of GISA and Abernethy's alg. are elaborated in this section. 
\cutsubsectionup
\subsection{Implementation of GISA}
\label{sec:implementationgisa}
\paragraph{Preference learning} Following Eq.~\eqref{eq:nll}, the posterior $p({\bf w};\mathcal{S})$ has the following form:
\begin{equation}
p({\bf w};\mathcal{S}^{(q)}) \propto \exp\left(-\frac{{\bf w}^T{\bf w}}{2C}\right) \prod_{q'=1}^{q} \frac{1}{1+\exp\left({\bf w}^T\Delta {\bf z}^{(q')}\right)}.
\label{eq:reglog}
\end{equation}
The maximum a posteriori (MAP) estimator is derived using the standard {\it liblinear} package~\cite{fan2008liblinear} with a ten-fold cross-validation for $C \in \{0.1,1,10,\cdots,10^8\}$ after each query. Once $C$ is determined, the resultant MAP estimator will be used to initialize the following MCMC procedure.

\begin{figure}
\centering
\includegraphics[width=\linewidth]{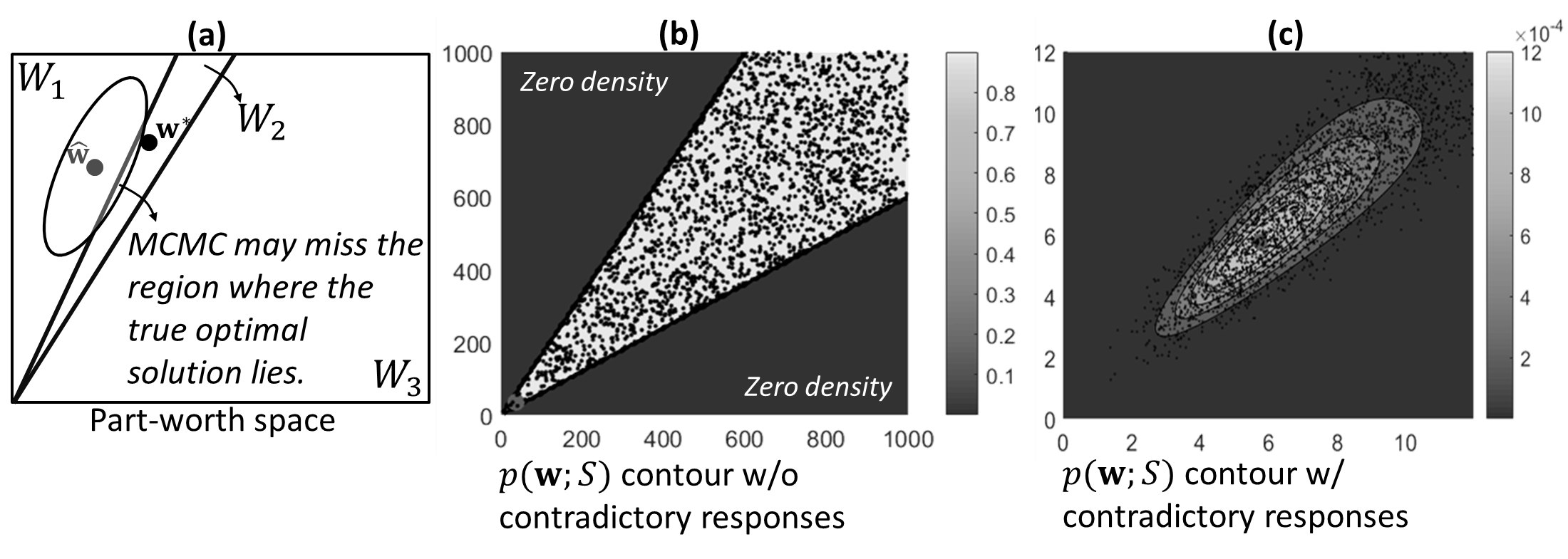}
\caption{(a) When $\mathcal{W}_{k^*}$ is narrow, naive MCMC has difficulty in identifying the true optimal design due to coarse numerical integrals. (b) Contour of $p({\bf w};\mathcal{S})$ in 2D. When responses are noise-free, the posterior distribution of ${\bf w}$ forms a plateau in the cone defined by $\Delta {\bf Z}$. (c) Contour of $p({\bf w};\mathcal{S})$ when contradictory responses exist. Generation of the contour: $\Delta {\bf Z}$ is composed of 10 rows of both $[1,-0.6]$ and $[-0.6,1]$, and one row of $[-1,0.6]$ and $[0.6,-1]$. The latter two represent choices against the true preference due to response noise.}
\label{fig:mcmc}
\end{figure}

\paragraph{Numerical integration through Markov chain Monte Carlo (MCMC)} Recall that executing GISA involves calculating $\pi_{l}$, $\pi_{l,k}$ and $\pi_k$ by integrating with respect to the density $p({\bf w};\mathcal{S})$. Here we discuss a MCMC approach to calculating these integrals in the absence of their analytical forms. We noticed during the development that the performance of GISA can be sensitive to the accuracy of these numerical integrals, 
and illustrate this issue using an artificial 2D case: Consider the three segments $\mathcal{W}_{1}$, $\mathcal{W}_{2}$, $\mathcal{W}_{3}$ in Fig.~\ref{fig:mcmc}a, among which $\mathcal{W}_2$ contains ${\bf w}^*$ and is ``narrow''. In this case, a good estimator $\hat{\bf w}$ can still fall into $\mathcal{W}_1$ or $\mathcal{W}_3$. With a coarse approximation, we could have $\pi_{2}^{(q)}=0$, and thus miss the true optimal design. Addressing this issue is non-trivial, particularly when responses are \highlight{close to} noise-free. To elaborate, Fig.~\ref{fig:mcmc}b shows an unnormalized density $p({\bf w};\mathcal{S})$ in a 2D case where $\mathcal{S}$ is composed of two vectors $[1,-0.6]$ and $[-0.6,1]$. The function forms a plateau between the two vectors, and only drops in value far away from the origin due to the prior $p({\bf w})$. Therefore, an accurate numerical integration requires samples to be drawn from the entire plateau. More generally, when responses are noise-free, $\mathcal{S}$ defines a convex cone $\{{\bf w}\in \mathbb{R}^D:\Delta {\bf Z}{\bf w}\geq {\bf 0}\}$ as the support of $p({\bf w};\mathcal{S})$, where the $q$th row of $\Delta {\bf Z}$ represents the difference $\Delta {\bf z}^{(q)} = {\bf z}^{(q,1)}-{\bf z}^{(q,2)}$. With a large number of queries, the cone becomes narrow, in which case both the standard Metropolis Hastings (MH)~\cite{chib1995understanding} or adaptive Metropolis~\cite{haario2001adaptive} algorithms produce a low acceptance rate, thus causing the aforementioned coarse approximation. \highlight{On the other hand, increasing the response noise level would cause contradictory responses to be observed more often, leading to a more smoothed density as can be seen in Fig.~\ref{fig:mcmc}c. The adaptive Metropolis algorithm can be applied in this case.}

\highlight{To this end, we introduce the following MH procedure that is applicable to all noise levels:}
\begin{algorithm}
\caption{Metropolis Hastings with a tailored proposal}\label{alg:mh}
\begin{algorithmic}[1]
\State {\bf Input}: $\Delta {\bf Z}$, $p({\bf w};\mathcal{S})$, $\hat{\bf w}$, sample size $J$
\State {\bf Output}: MH sample set $\mathcal{W}_{\text{MH}}$
\State set $\mathcal{W}_{\text{MH}}=\emptyset$, ${\bf w}_{\text{old}} = \hat{\bf w}$
\State delete rows of $\Delta {\bf Z}$ that have opposite signs, i.e., if $\Delta {\bf z}^{(q_1)} = -\Delta {\bf z}^{(q_2)}$ for some queries $q_1$ and $q_2$, these hyperplanes will be removed
\If {the remaining $\Delta {\bf Z}$ is not empty}
    \While {$j \leq 2J$}
    \State draw ${\bf d}$ uniformly from the unit sphere in $\mathbb{R}^D$
    \State calculate ${\bf a} := \Delta {\bf Z}{\bf d}$ and ${\bf b} := -\Delta {\bf Z}{\bf w}_{\text{old}}$
    \State define ${\bf a}_+$ and ${\bf b}_+$ (${\bf a}_-$ and ${\bf b}_-$) as a sub-vector of ${\bf a}$ and ${\bf b}$, respectively, where elements of ${\bf a}$ are positive (negative) 
    \State define $\delta_l := \max({\bf b}_+/{\bf a}_+)$ and $\delta_u := \min({\bf b}_-/{\bf a}_-)$, where $/$ is element-wise division
    \State if $\delta_l$ ($\delta_u$) does not exist, set $\delta_l = -10^3$ ($\delta_u = 10^3$)
    \State draw $\delta$ uniformly from $[\delta_l, \delta_u]$
    \State ${\bf w} = {\bf w}_{\text{old}} + \delta {\bf d}$
    \State calculate $\alpha = \min\left\{\frac{p({\bf w};\mathcal{S})}{p({\bf w}_{\text{old}};\mathcal{S})},1\right\}$
    \State draw $u$ uniformly from $[0,1]$
    \If {$\alpha > u$}
        \State $\mathcal{W}_{\text{MH}} \leftarrow {\bf w}$, ${\bf w}_{\text{old}} = {\bf w}$
    \Else
        \State $\mathcal{W}_{\text{MH}} \leftarrow {\bf w}_{\text{old}}$
    \EndIf
    \State $j = j+1$
    \EndWhile
\Else
    \State perform adaptive Metropolis according to \cite{haario2001adaptive}
\EndIf
\State return the second half of $\mathcal{W}_{\text{MH}}$
\end{algorithmic}
\end{algorithm}
Alg.~\ref{alg:mh} is modified from a standard MH for convex cones~\cite{boneh1979constraints}. The latter produces a sequence of samples, each of which perturbs from the current sample ${\bf w}_{\text{old}}$ by $\delta {\bf d}$, where ${\bf d}$ is a random unit vector, and $\delta$ is drawn uniformly from $[\delta_l, \delta_u]$ to ensure that the new sample, if accepted, is within the cone. Since the proposal density is a constant ($(\delta_u-\delta_l)^{-1}$), the acceptance level is simply $\alpha = \min\left\{\frac{p({\bf w};\mathcal{S})}{p({\bf w}_{\text{old}};\mathcal{S})},1\right\}$. Note that the cone is only guaranteed to be non-empty when responses are noise-free. To incorporate response noises, i.e., rows of $\Delta {\bf Z}$ with opposite signs, these rows are removed before running MH for the remaining convex cone. When the remaining cone is empty, the adaptive Metropolis algorithm will be applied. \highlight{In addition, the algorithm uses a single chain, and discards the first half of samples as burn-in samples. We do not adjust the algorithm for a target acceptance rate, unless adaptive Metropolis is used, when the target rate is $0.255$. Figs.~\ref{fig:mcmc}b,c demonstrate the performance of Alg.~\ref{alg:mh} under noise-free and noisy settings, respectively. Note that the scale of the contours in Figs.~\ref{fig:mcmc}b,c are less important since the density functions are unnormalized.}  

\paragraph{Choice of query candidates} In practice, the candidate query size $|\mathcal{Q}|$ can be large as it grows quadratically with respect to $|\mathcal{Z}|$. A large $|\mathcal{Q}|$ can make query selection costly. To this end, we propose to select a small set of $N$ candidate queries in each iteration using two heuristic criteria: Half of the candidates are chosen by first sorting designs with respect to $\Pi$ in descending order, and pairing designs starting from the top of this sorted list. E.g., let $\pi_{i_1}>\pi_{i_2}>\pi_{i_3}$. Then the candidate \highlight{pairs to be queried} are generated in the order of \highlight{${\bf z}_{i_1}$ and ${\bf z}_{i_2}$, ${\bf z}_{i_1}$ and ${\bf z}_{i_3}$, and lastly ${\bf z}_{i_2}$ and ${\bf z}_{i_3}$}. The second half are chosen based on Abernethy's ``minimax variance'' criterion (see Subsec.~\ref{sec:implementationaabernethy}). \highlight{Qualitatively, these criteria allow us to pick from queries that are either formed by highly probable designs, or those that will help to reduce the variance of $\hat{\bf w}$.} In addition, we omit the queries already chosen for the questionnaire.

\cutsubsectionup
\subsection{Implementation of Abernethy's algorithm.}
\label{sec:implementationaabernethy}
\paragraph{Choice of query candidates} In each iteration, $\hat{\bf w}$ and ${\bf H}(\hat{\bf w},\mathcal{S}^{(q)})$ are updated in the same way as in GISA\footnote{Note that Abernethy's original algorithm uses a regularization network (with squared loss and $l^2$ regularization on the estimates) for analytical part-worth estimation and thus has a different form of Hessian. It should be noted that due to the usage of a squared penalty (which allows the analytical solution), this formulation (see \cite{abernethy2008eliciting} for details) could falsely penalize an estimator that is consistent with query responses. The regularized logistic problem in Eq.~\eqref{eq:reglog}, on the other hand, does not have this issue and its MAP estimate can be derived by {\it liblinear} with a complexity of $O(Q)$, provided that the preference model is linear wrt ${\bf w}$. Thus we choose the latter formulation for estimation.}. The eigenvector ${\bf v}$ is calculated based on ${\bf H}_{\hat{\bf w}}$ (see Subsec.~\ref{subsec:benchmark}). To find such a query where $({\bf z}_{i}-{\bf z}_{j})$ is close to ${\bf v}$ and perpendicular to $\hat{\bf w}$, we calculate ${\bf c}_1 = |\Delta {\bf Z}\hat{\bf w}|$, where $\Delta {\bf Z}$ is a $|\mathcal{Q}|$-by-$d$ matrix, each row of which represents the difference between two designs of a candidate query in $\mathcal{Q}$; and ${\bf c}_2= |\Delta {\bf Z}{\bf v}|/||\Delta {\bf Z}||$, where $||\Delta {\bf Z}||$ is a $|\mathcal{Q}|$-by-$1$ vector with each element being the Euclidean norm of the corresponding row of $\Delta {\bf Z}$, and the division is element-wise. The query is then chosen by the following procedure: We first choose queries with the highest value in ${\bf c}_2$. If the resulting set contains more than one query, we choose a subset with the minimum value in ${\bf c}_1$. If multiple queries still remains, we pick the first one in the list. Similar to GISA, we remove previously chosen queries from $\mathcal{Q}$ during the questionnaire.

\cutsectionup
\section{Case study}
\label{sec:case}
We introduce a dial-readout scale design problem from Michalek et al.~\cite{michalek2005linking} to demonstrate the difference between GISA and Abernethy's alg.. 

\cutsubsectionup
\subsection{Design attributes and preference part-worths}
This problem considers six attributes that affect consumer
preference: weight capacity, aspect ratio, i.e., platform length divided by width, platform area, tick mark gap, i.e., distance between 1-lb tick marks, size of readout number and price. Discrete levels of these attributes are listed in Table \ref{tb:levels}. 
\begin{table}[ht!]
\centering
\caption{Design Attributes and Price Levels~\cite{michalek2005linking}}
\label{tb:levels}
\begin{tabular}{ l l c c c c c }
\hline 
Description & Units &  \multicolumn{5}{c}{Levels} \\
\hline
Weight Capacity & lbs & 200 &
250 & 300 & 350 & 400 \\
Aspect Ratio & - & 6/8 & 7/8 & 8/8 &
8/7 & 8/6 \\
Platform Area & $\text{in.}^2$ & 100 & 110
& 120 & 130 & 140 \\
Tick Mark Gap & in. & 2/32 & 3/32 &
4/32 & 5/32 & 6/32 \\
Number Size & in. & 0.75 & 1.00 & 1.25 & 1.50
& 1.75 \\
Price & \$ & 10 & 15 & 20 & 25 & 30 \\
\hline
\end{tabular}
\end{table}
Thus each design can be encoded by a 24 dimensional binary vector (Subsec.~\ref{sec:setup} explains why it is not 30 dimensional). The ``true'' part-worths, denoted as ${\bf w}^*$, were derived through a CBC study in
\cite{michalek2005linking} and are presented in Table \ref{tb:part-worth}.
\begin{table}
\centering
\caption{True part-worth values~\cite{michalek2005linking}}
\label{tb:part-worth}
\begin{tabular}{ l c l c l c }
\hline
\multicolumn{2}{c}{Weight Capacity} & \multicolumn{2}{c}{Aspect Ratio} &
\multicolumn{2}{c}{Platform Area} \\
\hline
200 lbs. & -0.534 & 0.75 in. & -0.744 & 100 $\text{in.}^2$ & 0.015  \\
250 lbs. & 0.129 & 1.00 in. & -0.198 & 110 $\text{in.}^2$ & -0.098  \\
300 lbs. & 0.228 & 1.25 in. & 0.235 & 120 $\text{in.}^2$ & 0.049  \\
350 lbs. & 0.104 & 1.50 in. & 0.291 & 130 $\text{in.}^2$ & 0.047  \\
400 lbs. & 0.052 & 1.75 in. & 0.396 & 140 $\text{in.}^2$ & -0.033  \\
\hline
\multicolumn{2}{c}{Tick Mark Gap} & \multicolumn{2}{c}{Number Size} &
\multicolumn{2}{c}{Price}
\\
\hline
2/32 in. & -0.366 & 0.75 & -0.058 &  \$10 & 0.719 \\
3/32 in. & -0.164 & 0.88 & 0.253 &  \$15 & 0.482 \\
4/32 in. & 0.215 & 1.00 & 0.278 &  \$20 & 0.054 \\
5/32 in. & 0.194 & 1.14 & -0.025 &  \$25 & -0.368 \\
6/32 in. & 0.100 & 1.33 & -0.467 &  \$30 & -0.908 \\
\hline
\end{tabular}
\end{table}

\cutsubsectionup
\subsection{Candidate design set}
The engineering model \highlight{from \cite{michalek2005linking}} maps some fourteen design variables, denoted as ${\bf x}$, to the attributes\footnote{``Variables'' are parameters that the designer has control of (e.g., engine size), while ``attributes'' are performance measures resulting from the variables, e.g., acceleration time of the car.}, and specifies constraints on the variables. Specifically, let
${\bf z}({\bf x})$ be the mapping from design variables ${\bf x}$ to attributes
${\bf z}$, and ${\bf g}({\bf x}) \leq {\bf 0}$ \highlight{be the engineering} constraints. To find the set of feasible designs in the attribute space, we solve the following problem:
\cutequationup
\begin{equation}
\begin{aligned}
& \min_{\bf x} && ||{\bf z} - {\bf z}({\bf x})||_2 \\
& \text{subject to} && {\bf g}({\bf x}) \leq {\bf 0}. \\
\end{aligned}
\cutequationdown
\end{equation}
A design ${\bf z}$ is considered feasible if a solution ${\bf x}$ is found and the resulting minimal discrepancy $||{\bf z} - {\bf z}({\bf x})||_2$ is less than a tolerance of $10^{-3}$. This procedure creates a set $\mathcal{Z}$ with $K=2455$ feasible designs, $5^6 = 15625$ out of the possible combinations.

\cutsubsectionup
\subsection{Simulation setup}
\label{sec:setup}
For both GISA and Abernethy's alg., the simulated questionnaire is set up as follows. We initialize by calculating the true optimal design ${\bf z}_{k^*}$ (Eq.~\eqref{eq:marketobj}) using $\theta{\bf w}^*$. We set $\theta=100$ to simulate noise-free responses: In this case, the false response rate $\text{sigmoid}(\theta\Delta u)$, for a utility gap $\Delta u:={\bf w}^T({\bf z}_i-{\bf z}_j)$ between two designs, is less than $0.01\%$ when $\Delta u=0.1$, i.e., consumers have a chance of $0.01\%$ to choose a product with less utility from a pair when the utility difference between the two is $0.1$. 

Note that part-worth estimation may encounter an identifiability issue: For any $\hat{\bf w}$ that optimizes Eq.~\eqref{eq:reglog} with a large enough $C$, in which case the regularization term can be neglected, an arbitrary part-worth vector $\hat{\bf w}+\delta_{\bf w}$ is also a solution, provided that $\delta_{\bf w}$ has the same part-worth values for attribute levels of each attribute. While the regularization term can theoretically address this issue when $C$ is small, the value of $C$ is typically large in the noise-free case. This is because through cross-validation, the algorithm learns to fully trust the responses rather than the prior. Therefore, we constrain $\hat{\bf w}$ by forcing the part-worth of the 5th level of each attribute, and thus $\delta {\bf w}$, to be zero. This treatment requires a slightly modification to the true part-worths ${\bf w}^*$ from Table~\ref{tb:part-worth}, by also shifting part-worth values of each attribute so that value of the 5th level of each attribute is zero\footnote{Another common practice of addressing the identifiability issue is to constrain part-worths for each attribute to a constant, e.g., $0$. However, doing so will introduce equality constraints to the estimation problem which are not naturally supported by the {\it liblinear} solver.}. This treatment leaves the part-worth space to be $\mathbb{R}^{24}$.

The candidate query set $\mathcal{Q}$ is set to contain all design pairs from $\mathcal{Z}$ with \highlight{$2455(2455-1)/2 = 3012285$ queries}. The first query of the questionnaire is randomly picked from $\mathcal{Q}$. For a query with designs ${\bf z}_{i}$ and ${\bf z}_{j}$, the simulated user chooses ${\bf z}_i$ with probability $\text{sigmoid}\left(\theta({\bf z}_{i}-{\bf z}_{j})^T{\bf w}^*\right)$. The lumped preference $u_0$ (for calculating the market share) is set by uniformly picking a \highlight{competing product} from $\mathcal{Z}$ during the initialization. To compare the performance of GISA and Abernethy's, we conduct $T=20$ questionnaires using both algorithms, each consisting of $Q=100$ queries. For GISA, we set the MCMC sample size to $J=10^3$ and \highlight{the size of the heuristic candidate set to be $N=100$}.

\cutsubsectionup
\subsection{Results and analysis}
\label{sec:results}
\paragraph{Performance measures} To compare the performance of GISA and Abernethy's alg., four measures are tracked along the questionnaire: (1) The probability $\pi_{k^*}^{(q)}$ of the true optimal design being identified as optimal, (2) the percentage of correct guesses across all independent runs: $\text{correct\%} := \frac{1}{T}\sum_{t=1}^T\mathbbm{1}\left(\pi_{k^*}^{(q)} >
\pi_{k}^{(q)},~\forall k \neq k^*\right)$, (3) the correlation $\hat{c} = \text{corr}(\hat{\bf w},{\bf w}^*)$, and (4) the estimation error $\hat{d} = ||\hat{\bf w}-{\bf w}^*||$. An algorithm is more effective at identifying the true optimal design if it has higher $\pi_{k^*}^{(q)}$ and $\text{correct\%}$ along the progress. On the other hand, the algorithm is better at identifying the true part-worths if $\hat{d}$ and $\hat{c}$ converge faster.

\begin{figure}
\centering
\includegraphics[width=\linewidth]{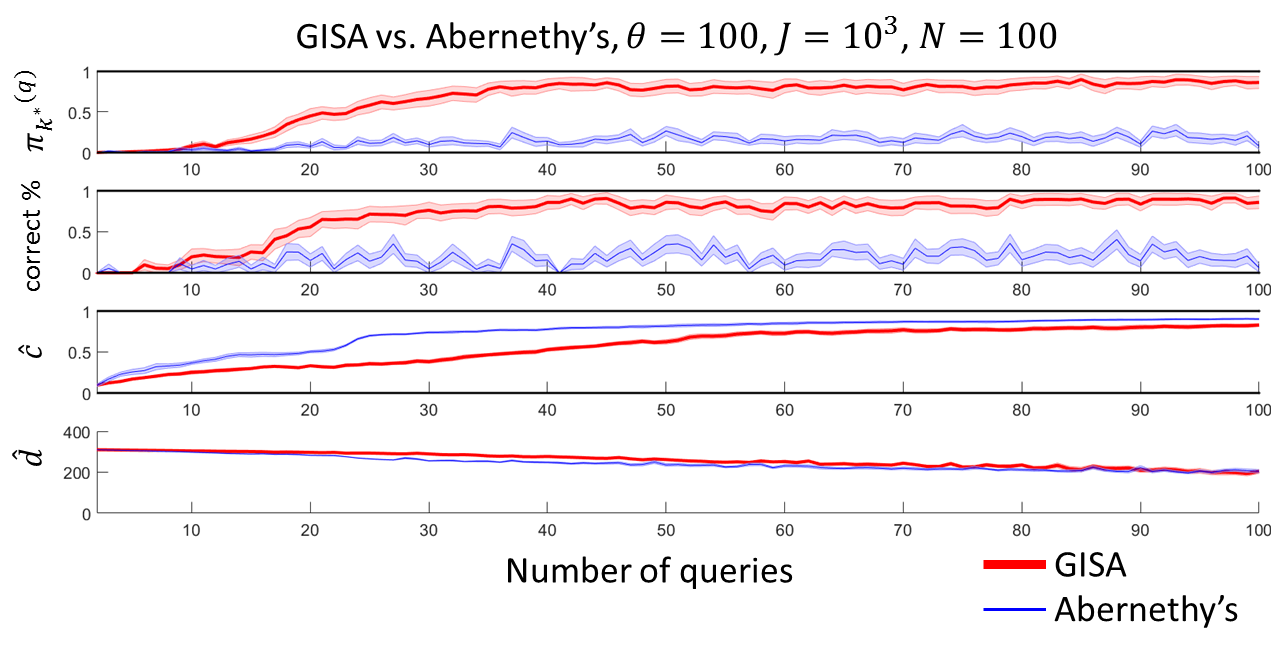}
\caption{Comparison between GISA and Abernethy's alg. under $\theta=100$, $J=10^3$, and $N=100$. The standard error of the mean is calculated based on 100 bootstrap samples. Best viewed in color.}
\label{fig:theta100}
\end{figure}

\paragraph{Main findings} Results from the case study are presented in Fig.~\ref{fig:theta100}: Under the noise-free condition, GISA outperforms Abernethy's alg. at identifying the optimal design while Abernethy's has better part-worth estimation with small $Q$. 

We also notice that the norm of $\hat{\bf w}$ does not converge to that of ${\bf w}^*$ in either case. This is because of the plateau in $p({\bf w}|\mathcal{S})$ under the noise-free assumption allows a variety of $\hat{\bf w}$ to be almost equally likely (see Fig.~\ref{fig:mcmc}b). \highlight{This is conceptually the same issue as characterizing the distribution of rare events. Note that $\hat{\bf w}$ will be close to the origin when response noise exists, and this issue will be alleviated. For completeness, we also note that on average $56\%$ of the queries are chosen from the first half of the heuristic candidate set (based on $\Pi$), and the rest from the ``minimax variance'' criterion. The average initial and final entropy values are $H(\Pi^{(0)}) = 8.22$ and $H(\Pi^{(100)}) = 0.14$. The initial entropy is lower than the theoretical maximum ($\log_2{2455}\approx 11$) since products with higher prices are intrinsically more likely to be optimal under a normal prior.}

\begin{figure}
\centering
\includegraphics[width=\linewidth]{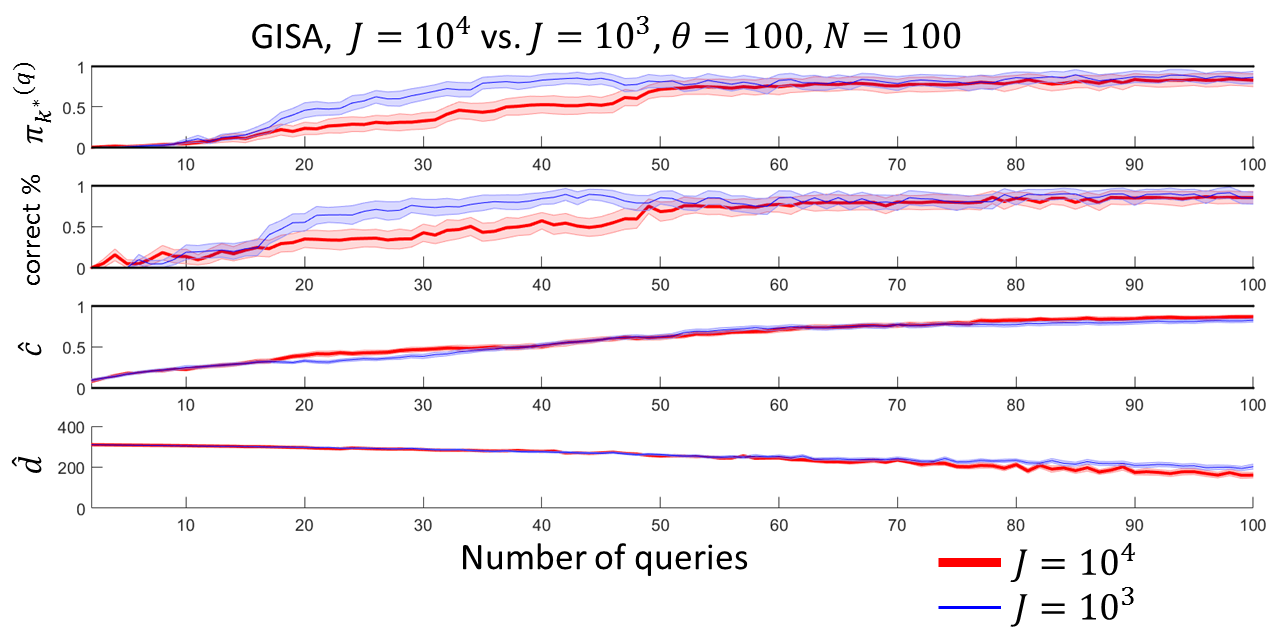}
\caption{Comparison between $J=10^4$ and $J=10^3$ under GISA, $\theta=100$, and $N=100$. Best viewed in color.}
\label{fig:figure3}
\end{figure}

\paragraph{Effects of algorithmic parameters} To further understand the influence of $J$ and $N$ on GISA's performance, we run the algorithm under $(J=10^4, N=100)$ and $(J=10^3, N=2)$ with $T=20$ and $Q=100$. \highlight{Based on our GISA implementation, the case with $N=2$ represents choosing between two candidate queries based on Eq.~\eqref{eq:obj}: One has two designs with the highest probabilities, and the other is the choice by Abernethy's alg..} Fig.~\ref{fig:figure3} summarizes the comparison between $J=10^4$ and $J=10^3$. \highlight{The result shows that an increased MCMC sample size could lead to lower convergence in $\pi_{k^*}^{(q)}$. This is because the large sample set may have a better chance to cover segments, especially the small ones, within the cone.} 
\begin{figure}
\centering
\includegraphics[width=\linewidth]{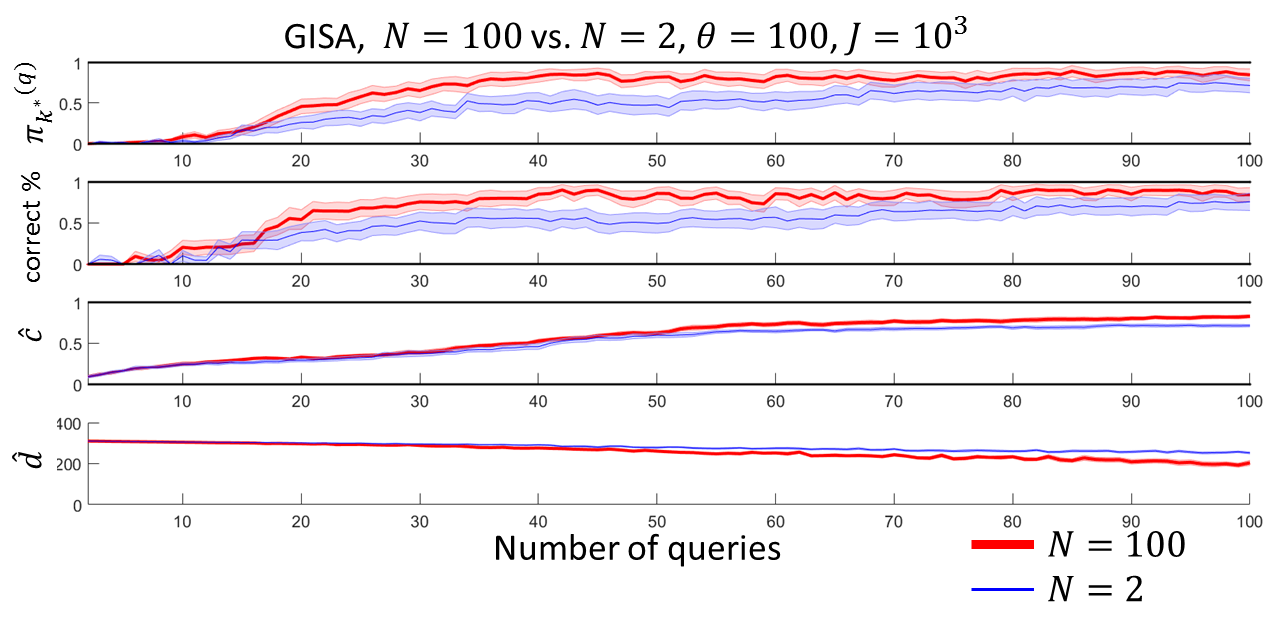}
\caption{Comparison between $N=100$ and $N=2$ under GISA, $\theta=100$, and $J=10^4$. Best viewed in color.}
\label{fig:figure4}
\end{figure}
\highlight{Fig.~\ref{fig:figure4} summarizes the comparison between $N=100$ and $N=2$.} By investigating results from individual simulations, we find that $N=100$ significantly outperforms $N=2$ \highlight{only} when $\mathcal{W}_{k^*}$ is narrow. This indicates that a larger set of queries, i.e., ways to cut the part-worth space, will allow GISA to perform more effectively. For concreteness, we report details of one particular case: Let ${\bf z}_{1}$, ${\bf z}_{2}$, ${\bf z}_{3}$ be designs with attribute levels listed in Table~\ref{tb:narrowcase}. Under the settings from Sec.~\ref{sec:case}, when ${\bf z}_3$ is picked as the single competing product, ${\bf z}_1$ is the true optimal design. Fig.~\ref{fig:figure6}b shows that one may identify an incorrect optimal design (${\bf z}_{2}$) when the part-worth estimator is slightly deviated from ${\bf w}^*$ in the dimensions for which the red lines are close to the black blocks. This shows that $\mathcal{W}_{k^*}$ is indeed narrow in this situation. Fig.~\ref{fig:figure6}a compares the performance of GISA with $N=100$ and $N=2$ under this particular case, with $T=20$ and $Q=200$. 
\begin{table}[ht!]
\centering
\caption{A case with narrow $\mathcal{W}_{k^*}$}
\label{tb:narrowcase}
\begin{tabular}{ l | c c c }
\hline 
Design & ${\bf z}_1$ & ${\bf z}_2$ & ${\bf z}_3$ \\
\hline
Weight Capacity & 300 & 300 & 250 \\
Aspect Ratio & 8/8 & 8/8 & 7/8 \\
Platform Area & 120 & 120 & 140 \\
Tick Mark Gap & 5/32 & 5/32 & 3/32 \\
Number Size & 1.25 & 1.25 & 0.75 \\
Price & 25 & 30 & 10 \\
\hline
\end{tabular}
\end{table}
\begin{figure}
\centering
\includegraphics[width=\linewidth]{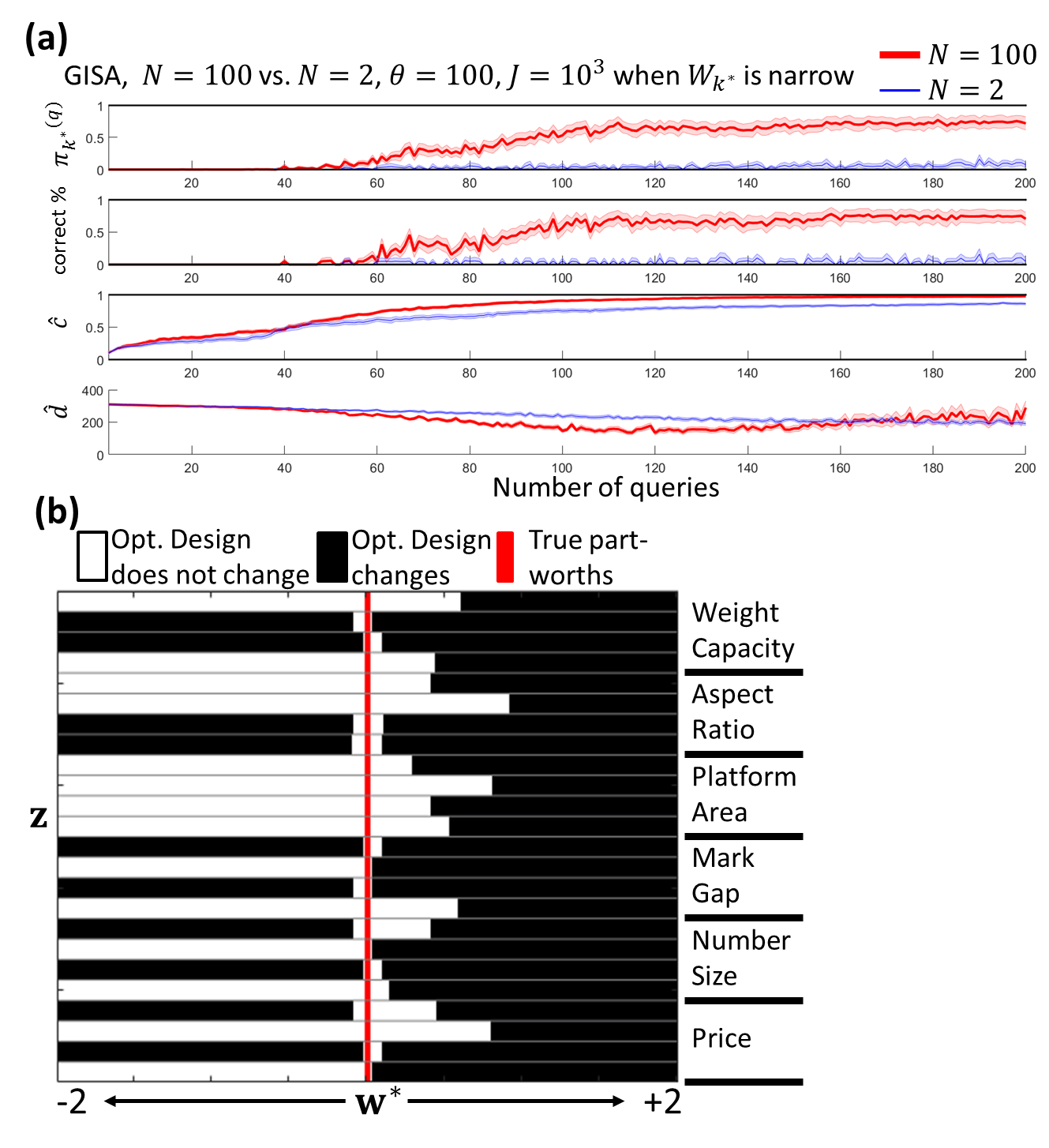}
\caption{(a) Comparison between GISA performance with $N=100$ and $N=2$ under the case in Table~\ref{tb:narrowcase} ($\theta=100$, $J=10^3$). (b) An illustration of how small deviations from ${\bf w}^*$ will change the optimal design when $\mathcal{W}_{k^*}$ is narrow: Each row represents part-worth values of an attribute level. Best viewed in color.}
\label{fig:figure6}
\end{figure}

\cutsubsectionup
\subsection{Influence of response noise (or preference indifference)}
\label{subsec:noise}
\begin{figure}
\centering
\includegraphics[width=\linewidth]{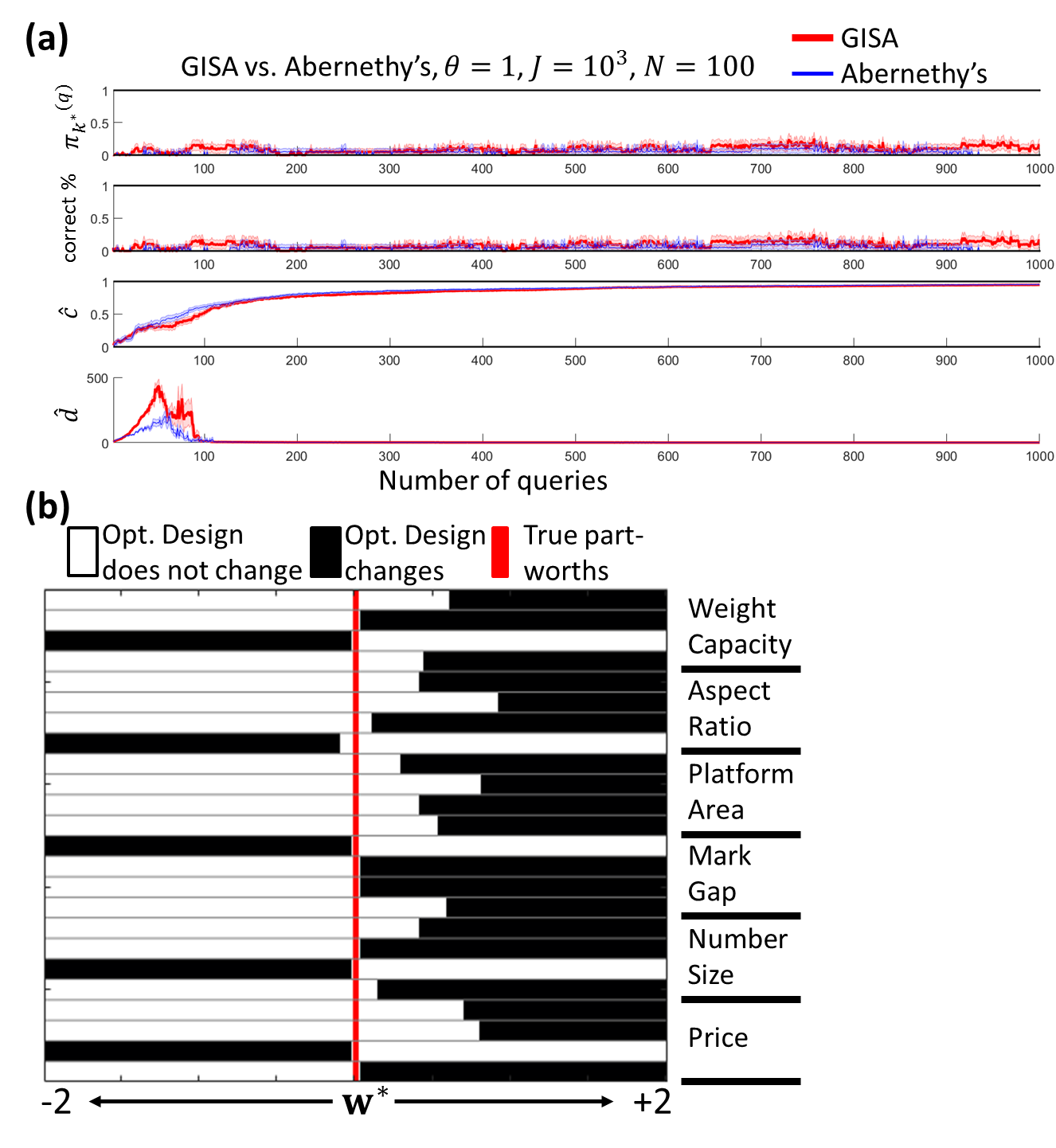}
\caption{(a) Comparison between GISA and Abernethy's alg. under $\theta=1$, $J=10^3$, and $N=100$. (b) An illustration of how small deviations from ${\bf w}^*$ will change the optimal design in the case of $\theta=1$: Each row represents part-worth values of an attribute level. Best viewed in color.}
\label{fig:theta1}
\end{figure}

\begin{figure}
\centering
\includegraphics[width=\linewidth]{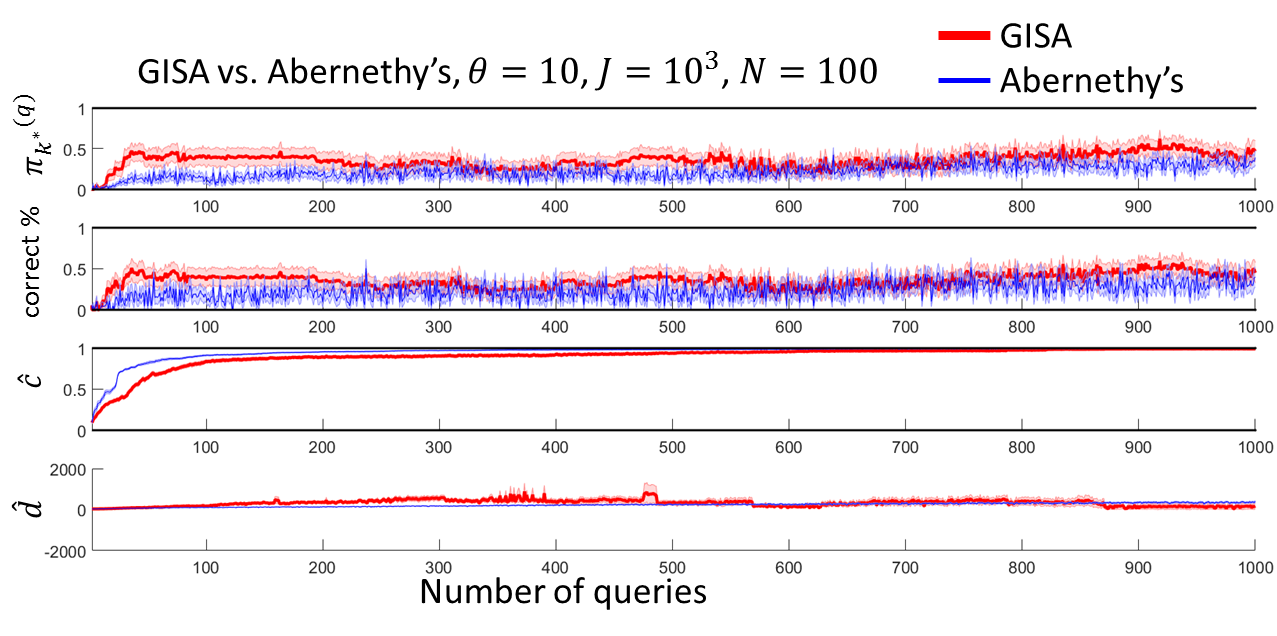}
\caption{(a) Comparison between GISA and Abernethy's alg. under $\theta=10$, $J=10^3$, and $N=100$. (b) An illustration of how small deviations from ${\bf w}^*$ will change the optimal design in the case of $\theta=1$: Each row represents part-worth values of an attribute level. Best viewed in color.}
\label{fig:theta10}
\end{figure}

We show in Fig.~\ref{fig:theta1}a the comparison under $\theta=1$, $J=10^3$, and $N=1000$. The false response rate in this case is $48\%$ for a utility gap of $0.1$. The result shows that we can reach a good part-worth estimation, as $\hat{c}$ and $\hat{d}$ can converge in both GISA and Abernethy's alg. (which confirms that the algorithms are robust for homogeneous and noisy part-worth estimation), but neither can correctly identify the optimal design. Recall the discussion in Sec.~\ref{sec:motivation}, the failure here is due to the clustering of segments in the part-worth space close to the origin. One example of this situation is visualized in Fig.~\ref{fig:theta1}b, which shows the narrow space of $\mathcal{W}_{k^*}$ around ${\bf w}^*$. Under this situation, and with the high uncertainty of the part-worth estimation due to high noise, the probability mass will be shared among all product candidates corresponding to the segments within the cluster around $\mathcal{W}_{k^*}$. \highlight{With an intermediate noise level $\theta=10$ (false response rate $27\%$ for a utility gap of $0.1$), we still observe significant performance drop from GISA. See Fig.~\ref{fig:theta10}.} Therefore, correct identification of the optimal design under noisy user responses is challenging by nature, as it requires both high accuracy in the point estimator and low uncertainty in the estimation. 


\begin{figure}
\centering
\includegraphics[width=\linewidth]{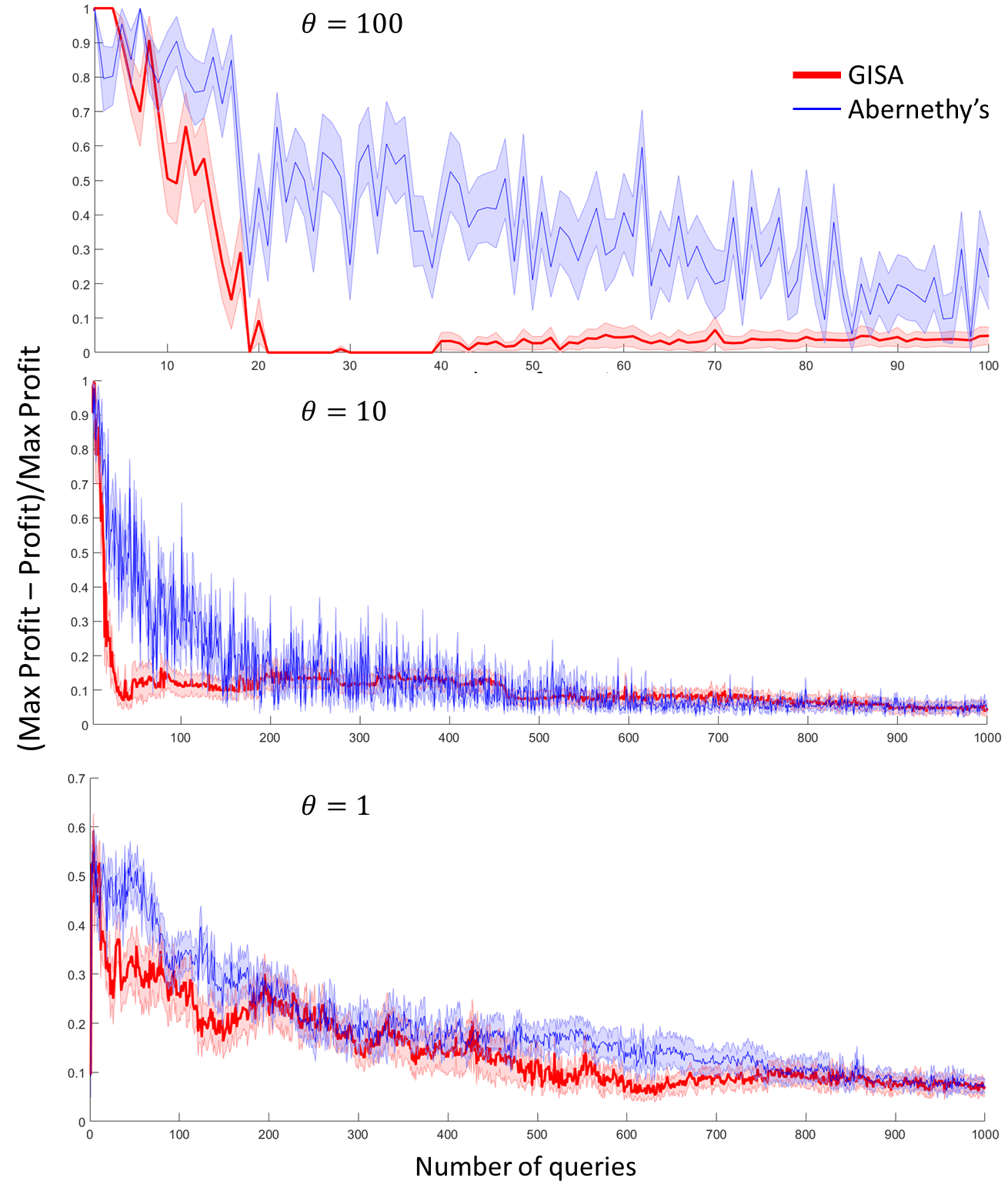}
\caption{\highlight{Profit gaps from GISA and Abernethy's alg. under $\theta=1, 10, 100$, $J=10^3$. The standard error of the mean is calculated based on 100 bootstrap samples. Best viewed in color.}}
\label{fig:profitthetacomp}
\end{figure}

\highlight{While GISA suffers from noise responses, the comparison among the three noise levels (Fig.~\ref{fig:profitthetacomp}) shows that the product chosen through GISA consistently has a higher profit than that through Abernethy's alg., with its advantage decreases with increasing noise.   
This finding highlights the necessity of GISA, as it shows that for two part-worth estimates close to the ground truth in correlation (and thus they could belong to two neighbouring segments in the part-worth space), the difference between their profits can be significant, especially when noise response is low. We shall also note that the noisy preference causes the profit variance among all products to be less significant, which allows both GISA and Abernethy's alg. to achieve a profit close to the true optimum when an adequate number of queries are made, even when $\theta=1$.}  

\cutsectionup
\section{Discussion}
\label{sec:disc}
\begin{figure}
\centering
\includegraphics[width=\linewidth]{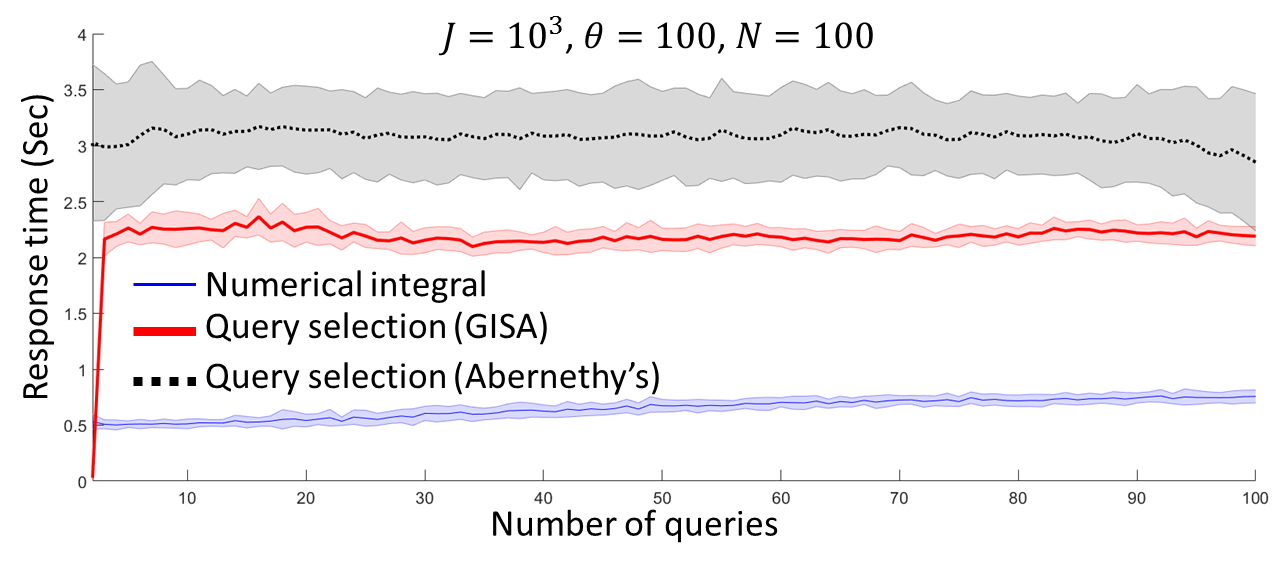}
\caption{Query response time for GISA and Abernethy's alg. under the presented case ($2455$ designs in $\mathbb{R}^{24}$).}
\label{fig:time}
\end{figure}

\paragraph{Applicability to real-time survey}
\label{subsec:realtime}
Fig.~\ref{fig:time} records the response time of GISA and Abernethy's alg. when executed on a workstation with Intel Xeon CPU at 2.10GHz, with a sample size of $J=10^3$, a candidate query size of $N=100$, a candidate design size of $K=2455$, and a total query size of $3012285$. The query selection steps for Abernethy's alg. is slightly slower than that of GISA as our implementation requires sorting of two arrays, both starting with $3012285$ elements, while GISA requires one. The cost of numerical integrals in GISA increases linearly with the size of $\mathcal{S}$. We also note that the computational complexity of estimating $\Pi$ scales linearly with the number of candidate designs. While GISA could thus become impractical if we consider candidate designs induced by a large set of attributes, such situations rarely happen in questionnaires as human beings can only consider a handful of attributes at a time.

\paragraph{Heterogeneous preferences and alternative preference models}
\label{subsec:relaxw}
Recall that GISA only requires $p({\bf w};\mathcal{S})$ for choosing a query, thus the algorithm can be extended for heterogeneous user preferences. To do so, let the true individual part-worths be drawn from a distribution $p({\bf w};\boldsymbol{\beta})$ with unknown parameters $\boldsymbol{\beta}$. For example, $\boldsymbol{\beta}$ can represent the mean and variance-covariance matrix of ${\bf w}$, or hyperparameters of distributions where these two are drawn from. We will then redefine $p({\bf w};\mathcal{S}):= \mathbb{E}_{\boldsymbol{\beta}}p({\bf w}|\boldsymbol{\beta})$ with respect to the learned posterior $p(\boldsymbol{\beta};\mathcal{S})$. In addition, other preference models, e.g., probit and non-compensatory models, can also be incorporated, provided that a function proportional to the posterior distribution $p({\bf w};\mathcal{S})$ can be calculated (${\bf w}$ will be a binary vector in the non-compensatory case~\cite{hauser2010disjunctions}). Nonetheless, such extensions may significantly increase the computation cost for calculating $\hat{\bf w}$ and the numerical integrals involved in GISA, \highlight{due to the additional integral involved in probit models and the combinatorial nature of estimating ${\bf w}$ of a non-compensatory model}.   

\paragraph{Optimal query}
\label{subsec:optquery}
Recall that the current implementation picks candidate queries heuristically to make the algorithm tractable. Theoretically, one can minimize the local cost $L^{(q)}$ with respect to the query without limiting feasible solutions to $\mathcal{Q}$, so that queries can be ``designed'' rather than chosen from the set. The cost-effectiveness of such an approach is yet to be studied.

\paragraph{Beyond binary queries}
\label{subsec:outside}
Our questionnaire assumes no ``outside good'', i.e., a user has to make a decision between the two presented designs. This can be relaxed by allowing the user to choose ``not to buy any of these'' and/or ``these are equally good''. The former requires the estimation of the utility of ``not choosing any''; and the latter sets a constraint on the utility gap between the two designs. The functional form of the posterior distribution $p({\bf w}|\mathcal{S})$ can be modified to incorporate both responses. In addition, it is feasible to extend the query to include more than two designs, in which case the user can choose more than one design as the more preferred ones. Such choice data can be interpreted as multiple binary choices and thus can be directly incorporated into the regularized logistic model. One caveat in doing so is that when random errors in the choice of multiple products during one query is correlated, a probit model will be used in place of multiple logit ones, which makes the estimation much harder.

\paragraph{Product line design}
\label{subsec:productline}
While this paper focused on the design of a single product, optimal product line design can be handled by considering each possible combination of products as a meta-design, i.e., during the questionnaire, we will calculate the probability for product lines to be the most profitable, instead of individual designs. The computational cost for query selection increases linearly with the number of candidates (combinations of designs). Therefore with the combinatorial nature of product lines, this problem could still be intractable. Nonetheless, such an issue is universal to product line design, regardless of the choice of a query strategy. The added value of this paper is that it enables a more effective query when one can afford it. We shall also note that the cost of numerical integration will not be affected as the dimensionality of the part-worth space does not change.

\paragraph{Alternative definition of optimality}
\label{subsec:defopt}
Recall that in Subsec.~\ref{sec:optid} we proposed two definitions of optimality under the existence of uncertainty in part-worth estimation: The design with the highest probability to be the most profitable, and that with the highest expected profit. These two are different in general: The former may have less expected profit, while the latter may have higher variance in profit. Nonetheless, when we reach $\pi_{k^*} = 1$, the optimal design ${\bf z}_{k^*}$ is the solution under both definitions. \highlight{However, we have not yet investigated which of these two measures will lead to faster correct identification of the optimal product design, and in what conditions.}

\paragraph{Infinite number of candidate products}
\label{subsec:infinite}
While the proposed query method is developed for a finite candidate set, it is sometimes desired that the optimal product can be found from a continuous space. To take dial-readout scale case as an example, once we identified that a price at \$10 is the optimal setting among all five levels, one would like to know if \$9 or \$11 would lead to even better profit. This refined identification problem can be solved by continuing the query process with the current $p({\bf w}|\mathcal{S})$ and a new segmentation of the part-worth space produced by the refined candidate products.  

\paragraph{Relevance to value-based global optimization (VGO)}
\label{subsec:vdd}
VGO~\cite{moore2014value} is the process of solving black-box optimization problems through an adaptive sampling scheme where each sample is chosen to maximize the ``value of information'', i.e., the expected difference between the potential profit after one more sample (minus the cost of the sample) and the current profit. VGO is similar to the proposed method in that it uses a local objective to guide the query. However, the value of information measure is not derived directly from solving Eq.~\eqref{eq:marketobj1} or Eq.~\eqref{eq:marketobj2}. In addition, the original formulation of VGO assumes query responses to be evaluations of the objective (e.g., profit), whereas in this paper, we focused on queries that are pair-wise comparisons of a measure (preference) that can indirectly infer the objective. It would be interesting, nonetheless, to extend VGO to the setting of optimal product identification in an extension to this work.

\cutsectionup
\section{Conclusions}
\label{sec:conclusion}
While most research on design for market systems has focused on post-data-collection
scenarios where adequate data is assumed to be available for identifying the
optimal designs, this work investigated how cost-effective questionnaires shall
be adaptively constructed for the purpose of identifying the most profitable design from a finite candidate set. 

We showed that the two objectives, preference modeling and optimal product identification, require their own questionnaire mechanisms due to their difference in problem formulation. To summarize, the former prefers queries that quickly concentrate the part-worth distribution towards the true part-worths, while the latter requires queries to shift the distribution to a segment of the part-worth space that is defined by the true optimal design. \highlight{Based on this, we proposed an extension of a group identification algorithm (GISA) that directly minimizes the expected number of queries needed to identify the optimal product. While developed based on a noise-free model, GISA consistently outperformed a standard adaptive questionnaire algorithm with respect to both
the accuracy of identifying the true optimal design and the resultant expected profit. This consistency is shown through multiple independent runs of a simulated case study with randomized settings. The advantage of GISA, however, reduces with increasing response noise level. Our findings encourage deeper investigation into knowledge acquisition strategies that smartly seek for and integrate engineering and marketing data to facilitate low-cost and accurate identification of the optimal product design.}



\cutsectionup
\section*{Acknowledgement}
\label{sec:ack} The authors would like to thank Professor Fred Feinberg for
his advice and valuable comments on this work. This work has been supported by
the National Science Foundation under Grant No. CMMI-1266184. This support is
gratefully acknowledged.
\bibliographystyle{asmems4}
\bibliography{detc2014active}
\end{document}